\documentclass[pdflatex,sn-nature]{sn-jnl}% Style for submissions to Nature Portfolio journals
\usepackage{graphicx}%
\usepackage{multirow}%
\usepackage{amsmath,amssymb,amsfonts}%
\usepackage{amsthm}%
\usepackage{mathrsfs}%
\usepackage[title]{appendix}%
\usepackage{xcolor}%
\usepackage{textcomp}%
\usepackage{manyfoot}%
\usepackage{booktabs}%
\usepackage{algorithm}%
\usepackage{algorithmicx}%
\usepackage{algpseudocode}%
\usepackage{listings}%
\usepackage{booktabs,siunitx}
\begin{document}

\title[Article Title]{
  From Raw Segmentations to Simulation-Ready Cardiac Meshes: An Automated Framework for Anatomical Reconstruction and Virtual Cohort Generation
  
% From raw segmentations to computational meshes: an automated framework to create simulation-ready virtual cohorts of cardiac anatomies

% Create virtual cohorts of cardiac anatomies: an automated framework based morphing and statistical shape modelling

%Cardiac virtual cohorts improving raw segmentations using morphing

% Create virtual cohorts of simulation-ready cardiac anatomies: an automated framework based morphing and statistical shape modelling

% Simulation-ready virtual cohorts of cardiac anatomies: an automated framework based on morphing and statistical shape modeling

% Simulation-ready cohorts of cardiac anatomies: an automated framework based on morphing and statistical shape modeling

}
\author[1]{\fnm{Francesco} \sur{Fabbri}}
\author[1]{\fnm{Martino Andrea} \sur{Scarpolini}}
\author[2]{\fnm{Paolo} \sur{Ciancarella}}
\author[3,1]{\fnm{Francesco} \sur{Tudisco}}
\author[1,4]{\fnm{Roberto} \sur{Verzicco}}
\author[2]{\fnm{Alessandro} \sur{Ricci}}
\author*[1,5]{\fnm{Francesco} \sur{Viola}}\email{francesco.viola@gssi.it}

\affil[1]{\orgname{Gran Sasso Science Institute (GSSI)}, \orgaddress{\city{L'Aquila}, \country{Italy}}}

\affil[2]{\orgname{European Hospital}, \orgaddress{\city{Rome}, \country{Italy}}}

\affil[3]{\orgname{University of Edinburgh}, \orgaddress{\city{Edinburgh}, \country{Scotland}}}

\affil[4]{\orgname{Physics of Fluids Group, University of Twente}, \orgaddress{\city{Enschede}, \country{Netherlands}}}

\affil[5]{\orgname{INFN–Laboratori Nazionali del Gran Sasso}, \orgaddress{\city{Assergi}, \country{Italy}}}


\abstract{Computational models of the human heart are widely used to study electromechanical and fluid-dynamical cardiac function and to support applications such as in silico clinical trials. However, most studies remain limited to single or patient-specific anatomies, restricting the inclusion of population-level variability required for robust uncertainty quantification. A key challenge is translating medical-image segmentations, which may contain artifacts, mesh defects, holes, or disjoint domains, into topologically coherent geometries suitable for multiphysics simulations.

In this work, we present a semi-automatic pipeline that converts CT-based segmentations into simulation-ready cardiac meshes within a few minutes while preserving anatomical and topological consistency. Building on modern deep learning segmentation methods, the framework incorporates a template-based registration stage to regularize artifacts and enforce mesh-quality constraints. A Chamfer-distance-driven morphing strategy deforms a high-quality template toward each segmented heart, matching individual chambers to their anatomical counterparts while preserving topology and smooth transitions across chamber interfaces.

The resulting meshes are watertight, isotopological, and endowed with consistent point-to-point correspondence. The pipeline is validated on 58 healthy cardiac CT scans, including all cardiac chambers and proximal vessel segments. The resulting meshes can be represented in a unified shape space, enabling the construction of a statistical shape model of the heart and major vessels. Principal Component Analysis shows that a low-dimensional latent space efficiently captures population variability, while Gaussian Mixture Modeling enables synthetic anatomy generation. Overall, the proposed framework (whose implementation is released open-source)  provides a robust pathway from raw segmentations to simulation-ready cardiac geometries, enabling anatomically consistent virtual cohorts for large-scale in silico studies, uncertainty quantification, and analysis of variability-driven cardiac function.

}




\maketitle

\section{Introduction}

Computational modeling of the human heart is a key tool for investigating the electromechanical and fluid dynamical mechanisms underlying both physiological and pathological cardiac function. The ever-increasing availability of computational power has significantly expanded the interest in such multiphysics models across a wide range of scenarios. Among many, one of the most promising applications is the development of \emph{in-silico} clinical trials, which can support the design of cardiac prostheses and assist clinicians during interventions by providing high-fidelity quantitative predictions \cite{mcqueen2000three, takizawa2014estimation,10.1115/1.2746378,abbas2022state,VIOLA2020212,VIOLA2022108248,scarpolini_reg,formaggia2010cardiovascular,vedula2015hemodynamics,de2009direct,LUPI2026204445}. In this context, computational models are increasingly regarded as complementary tools to experimental and clinical studies, enabling controlled and repeatable analyses that are otherwise impractical or infeasible in-vivo.

The majority of existing \emph{in-silico} studies are limited to single, often patient-specific, anatomies. In contrast, population-level variability must be incorporated to enable robust numerical modeling, uncertainty quantification, and ultimately the realization of \emph{in-silico} clinical trials \cite{BOSNJAK2025110017, cheng2004three, 10.1093/pnasnexus/pgae392}. This requires access to virtual cohorts of anatomically consistent cardiac geometries. To this end, the primary source of information is medical imaging data, which poses significant challenges related to data acquisition, curation, and processing at scale. Furthermore, inter-subject variability in anatomy, imaging protocols, and resolution introduces additional sources of uncertainty that must be carefully accounted for when constructing representative cohorts.

Traditional segmentation approaches are often unable to produce robust and scalable anatomical cohorts due to acquisition and operator-dependent limitations \cite{ataei2021evaluation,covert2022intra,fischer2025quantifying}. In recent years, substantial progress has been achieved thanks to (i) large-scale efforts by healthcare institutions to systematically organize, curate, and share medical imaging datasets \cite{Sudlow2015UKBiobank,Pace2024HVSMR,10.1007/978-3-030-59719-1_8,4601463,stevens2020deep}, and (ii) the rapid advancement of deep learning (DL) methodologies for medical image analysis \cite{tsuneki2022deep,li2023medical,shen2017deep}, including state-of-the-art frameworks such as TotalSegmentator \cite{Wasserthal2023TotalSegmentator,doi:10.1148/radiol.241613,FEDOROV20121323} and nnU-Net \cite{Isensee2021nnUNet,Ise_nnUNet_MICCAI2024}. These approaches have enabled large-scale automated processing pipelines, significantly reducing the time and manual effort required to extract anatomical information from volumetric data.

\begin{figure*}[!htb]
    \centering
    \includegraphics[width=0.99\textwidth]{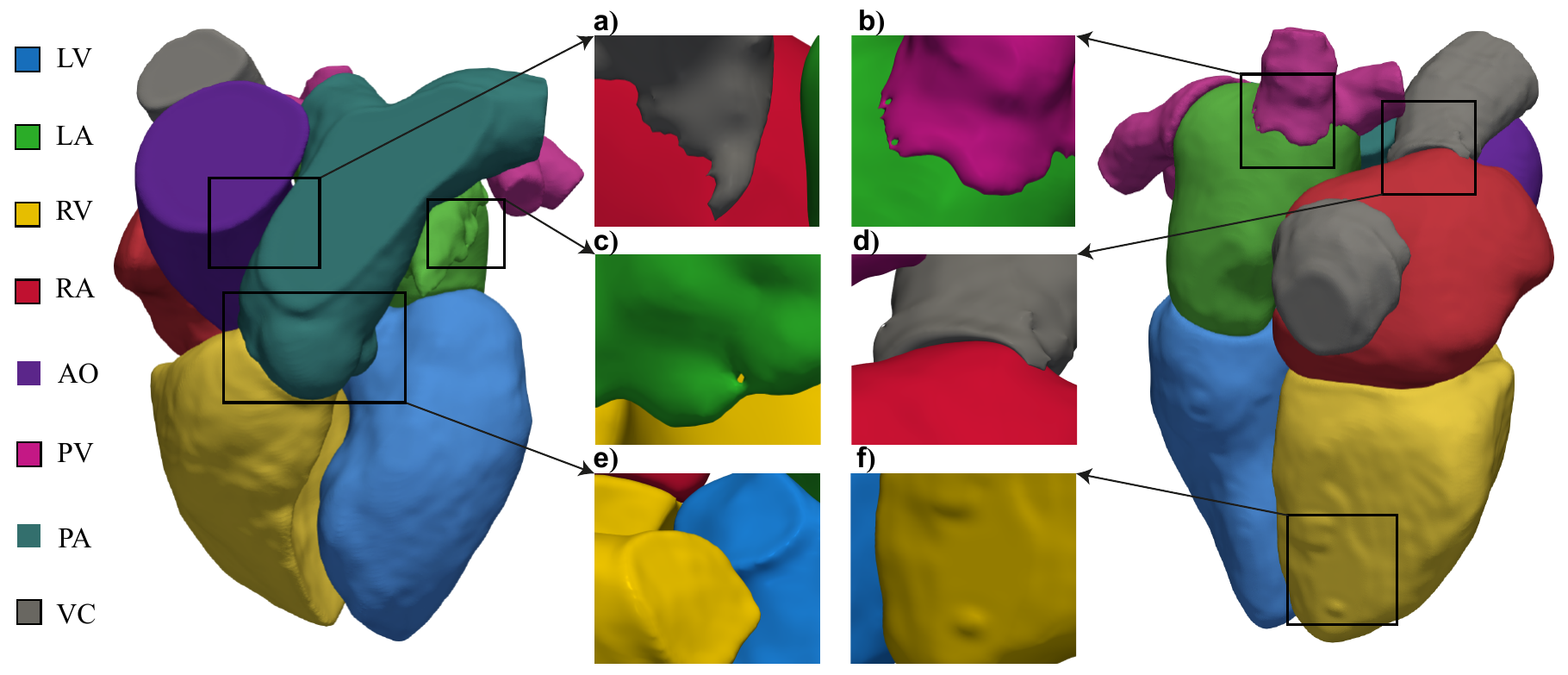}
    \caption{AI-based semi-automatic segmentation of the cardiac anatomy. Each anatomical structure is colored according to the legend on the left. The magnified panels highlight representative segmentation artifacts that can arise from automatic or semi-automatic segmentation. These include small holes and discontinuities at junctional regions, such as the connections involving the vena cava and pulmonary veins (panels a, d, and b, respectively). In addition, similar intensity values between neighbouring regions further complicate boundary detection, resulting in holes and local inconsistencies (such as in the left atrial appendage and surrounding tissues in panel c). Moreover, multi structure segmentation artificially close anatomical openings between neighbouring structures, preventing a physiologically meaningful representation of flow through regions such as the aortic and pulmonary valve planes (panel e). Finally, local mesh defects on chamber surfaces are further illustrated in panel (f). In selected insets, foreground structures were hidden to expose defects that would otherwise be occluded in the complete multi-chamber rendering.}
    \label{fig:defects}
\end{figure*}

Despite the above advances, DL-based methods, while enabling fast and fully automated segmentation of clinical data ($\sim$one minute), typically generate three-dimensional anatomical reconstructions primarily intended for visualization or diagnostic purposes. As a result, they may exhibit geometric and topological defects, such as holes, discontinuities, or incorrect connections between adjacent anatomical structures, particularly in junctional regions (e.g., vena cava, pulmonary veins, or valve areas; see Figure~\ref{fig:defects}). Such inconsistencies may arise from ambiguous image boundaries, limited resolution, or automated post-processing steps that alter anatomical openings. These defects prevent the direct use of segmented geometries in multiphysics simulations, especially in computational hemodynamics \cite{iglesias2015multi,panayides2020ai,pmlr-v222-zheng24a}, which require watertight and topologically consistent domains. Consequently, additional geometry processing and quality control steps are often required, which can be time-consuming and may introduce user-dependent variability, thereby limiting scalability. This gap between clinical imaging outputs and the requirements of high-fidelity computational modeling remains a major bottleneck.

These limitations stress the need to construct large-scale, high-quality datasets that explicitly enforce both anatomical accuracy and topological coherence, while leveraging existing heterogeneous patient-based datasets as sources of morphological variability for statistically meaningful analyses. In particular, there is a growing demand for methodologies that can bridge the gap between raw segmented data and simulation-ready geometries in a systematic and reproducible manner.

In this work, we propose a semi-automatic pipeline for generating simulation-ready meshes of the whole heart, including main vessels and arteries, directly from computed tomography (CT) scans. The method is applied to reconstruct the endocardium of the entire human heart, including the four heart chambers (two ventricles and two atria) and the initial tracts of the main arteries/vessels (aorta, pulmonary veins and artery, vena cava).
The proposed framework overcomes the limitations of traditional segmentation-based approaches through a template-based, differentiable, non-rigid multi-chamber registration approach. The latter is formulated as the minimization of a Chamfer distance functional and solved using a multi-scale optimization strategy that captures both global and local anatomical variations, yielding consistent, watertight meshes with shared topology. 
The resulting registered dataset of cardiac anatomies enables statistical shape modeling (SSM), providing a compact and parametric representation of anatomical variability while enforcing geometric consistency across populations. SSMs have been widely adopted in cardiac modeling for atlas construction, population analysis, and data-driven simulation studies \cite{ordas2007statistical,BAI2015133,rodero2021linking,KONG2024103293,scarpolini_reg}. Moreover, by enabling the generation of synthetic yet anatomically plausible geometries, SSMs play a crucial role in augmenting limited datasets and supporting uncertainty quantification in computational pipelines.

The pipeline integrates deep learning–based segmentation, principal component analysis, and in-house geometric processing libraries to ensure mesh quality compatible with multiphysics solvers. Applied to a dataset of 58 cardiac CT scans, the framework transforms heterogeneous segmentation outputs into coherent, simulation-ready heart models suitable for electrophysiological, mechanical, and fluid-dynamical simulations. Because all reconstructed geometries share a common topology, a statistical shape model of the heart can be constructed. Principal Component Analysis (PCA) provides a compact representation of morphological variability, while a Gaussian Mixture Model (GMM) captures the probability distribution of the reduced space, enabling meaningful quantitative analysis and the generation of anatomically plausible virtual cardiac cohorts.

\section{Results}\label{sec:result}
% This section presents the evaluation of the proposed framework, examining both geometric accuracy and statistical behavior across the dataset.
% The registration stage is first analyzed in \ref{sec:registr}, where the convergence properties of the template-refinement procedure are assessed through the evolution of the optimization loss. Anatomical consistency is subsequently investigated \ref{sec:result_mesh_a} by comparing morphogeometric descriptors between the anatomical  $\{\mathcal{M}^k\}_{k=1}^{m}$ and registered meshes $\{\mathcal{M}^k_r\}_{k=1}^{m}$.
% Finally, the statistical modelling component is examined \ref{modelred}, including dimensionality reduction behavior, reconstruction performance with respect to compactness and generalization of the model, clustering structure, and statistical properties derived from the PCA and Gaussian mixture representations.
%\label{sec:dataset}

\subsection{Dataset registration}\label{sec:registr}
\indent
The proposed framework is applied to the CT dataset introduced in section~\ref{sec:dataset}. During the registration phase, for each $i-th$ homogenized anatomy the loss function $\mathcal D^i(\mathcal{M}^i_{\mathcal{T}},\mathcal{M}^i_{\mathcal{H}})$ is minimized in order to deform the template mesh, $\mathcal{M}^i_{\mathcal{T}}$, onto $\mathcal{M}^i_{\mathcal{H}}$. The registration of each geometry is performed on a computing node running Ubuntu 22.04 LTS equipped with an Nvidia A100 GPU. Each registration requires approximately four minutes using the prescribed multi-resolution setup with 4,000 and 20,000 vertices (other minor steps of the registration procedure, such as the affine transformation, are computed instantly and do not add any overhead to the overall registration computation time). This represents a substantial acceleration of the mesh-generation workflow, reducing the time required to obtain a simulation-ready cardiac mesh from an AI-based semi-automatic segmentation from approximately one week to only a few minutes per patient. Overall, registering the full cohort of 58 patients across 4 template-update iterations took approximately 15 hours.

\begin{figure*}[!htb]
    \centering
    \includegraphics[width=0.99\linewidth]{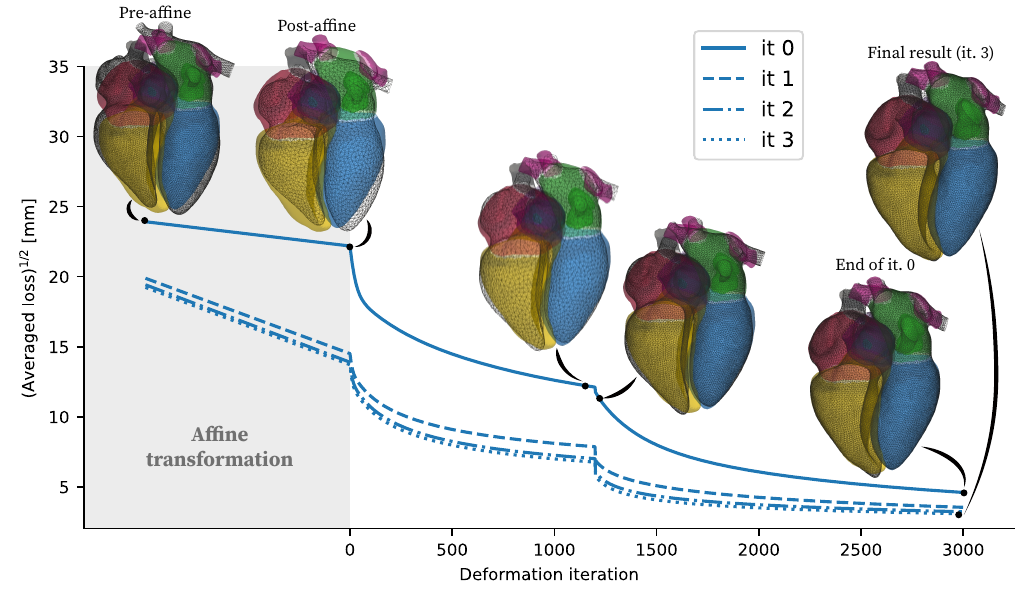}
    \caption{Loss evolution (root Chamfer distance of Equation~\ref{eq:cd_formal} averaged across the whole dataset) during the registration (both affine and deformation phases). The four curves correspond to the four iterations on the template mesh, each time picking the new average geometry as the new template. The overall loss decreases from $\sim 25~\text{mm}$ (iteration 0, pre-affine) to $3~\mathrm{mm}$ (final iteration on the template, post-deformation), reflecting improved alignment. The change of slope of the loss at epoch 1200 is due to a resolution change during the multi-scale nature of the registration algorithm. Snapshots of a randomly picked geometry are shown at specific instants during the registration.}
    \label{fig:loss}
  \end{figure*}

\begin{figure*}[!htb]
    \centering
    \includegraphics[width=0.99\linewidth]{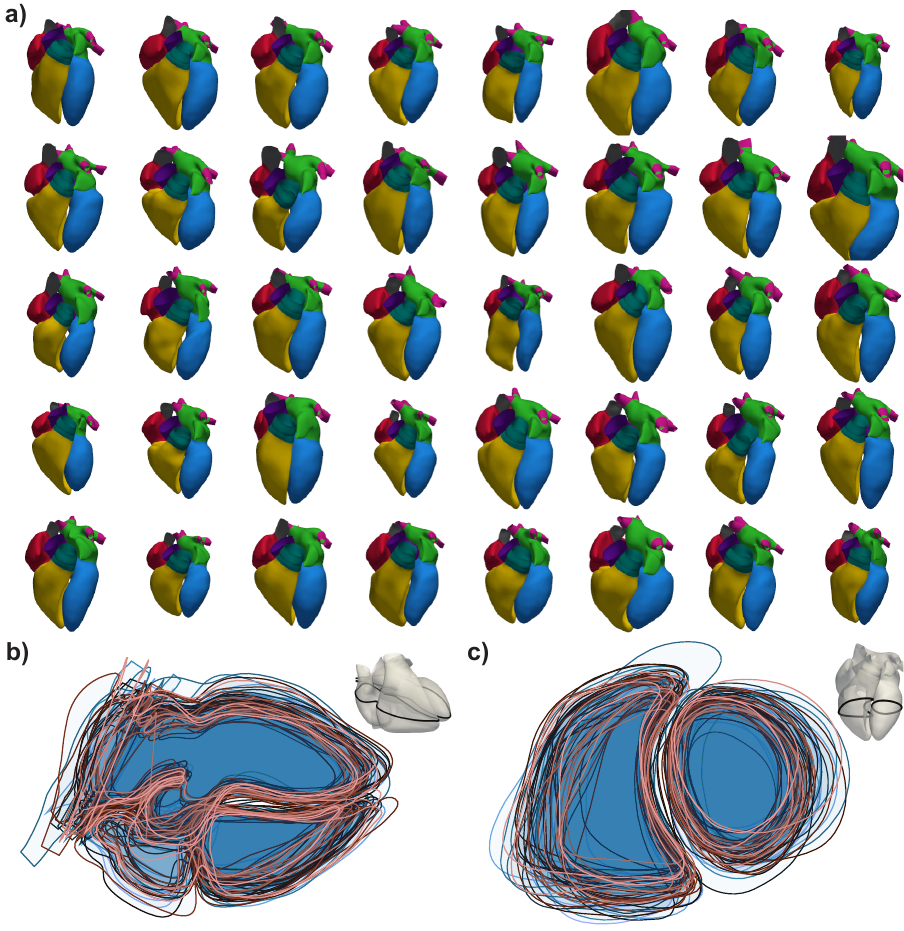}
    \caption{Visualization of the cohort obtained after template-to-anatomy registration. Panel (a) shows a subset of 40 meshes with identical topology and point-to-point correspondence, highlighting the consistency of the reconstructed cardiac structures. Panels (b) and (c) show the full cohort by superimposing slices from each geometry in the two-chamber and four-chamber views, respectively.}
    \label{fig:full_set}
\end{figure*}

The functional minimization during the gradient descent iterations averaged over the entire dataset, defined as $\sqrt{\sum_i \mathcal D^i}/m$ (with $m=58$ the number of homogenized anatomies), is shown in Figure~\ref{fig:loss}.
According to the proposed iterative procedure (see Figure~\ref{fig:big}), the higher curve corresponds to the registrations based on the initial template, whereas the lower ones indicate the registrations using the updated templates (defined by averaging the registrated dataset in the previous iteration, see section~\ref{sec:registration}), meaning that the updated template provides a closer initial geometric match to the population.
Since the loss function in the last two cases reaches the same minimum value (within a tolerance of 0.3~mm), the iterative registration procedure is terminated after the fourth loop.
The sharp decrease observed at epoch $1200$ corresponds to a scheduled transition in the multi-resolution optimization scheme, where the template is refined to a higher resolution mesh to capture finer geometric details (see section~\ref{sec:registration}).
\\ \indent
Figure~\ref{fig:full_set} depicts the resulting dataset of registered meshes. Panel (a) shows a subset of 40 meshes rendered from the same anterior camera position. Panels (b) and (c), instead, show superimposed cohort sections in two approximately orthogonal long-axis planes, corresponding to the four-chamber (b) and two-chamber (c) views, respectively. Visual inspection confirms that the registration procedure yields anatomically coherent geometries across the cohort while preserving subject-specific morphological variability. Several instances display variations in overall heart size, chamber elongation (longitudinal area), width (transversal area), curvature and torsion. The right ventricle, in particular, shows pronounced variability in shape and orientation, with some subjects exhibiting a more compact configuration and others presenting a visibly extended or laterally displaced structure. This variability is consistent with the complex geometry of the right ventricle and indicates that the deformation model does not enforce artificial geometric constraints.
No evident distortions or unrealistic shape artifacts are observed; chamber and vessels boundaries remain smooth and anatomically consistent across subjects, with no visible discontinuities, as highlighted by the chamber-wise coloring, which emphasizes the preservation of regular interfaces between adjacent regions and reflects the shared template topology.
A quantitative assessment of geometric fidelity and morphogeometric retention is presented in the following subsection.

\subsection{Anatomical validation}\label{sec:result_mesh_a}
\begin{figure}[htb!]
  \centering
    \includegraphics[width=0.95\linewidth]{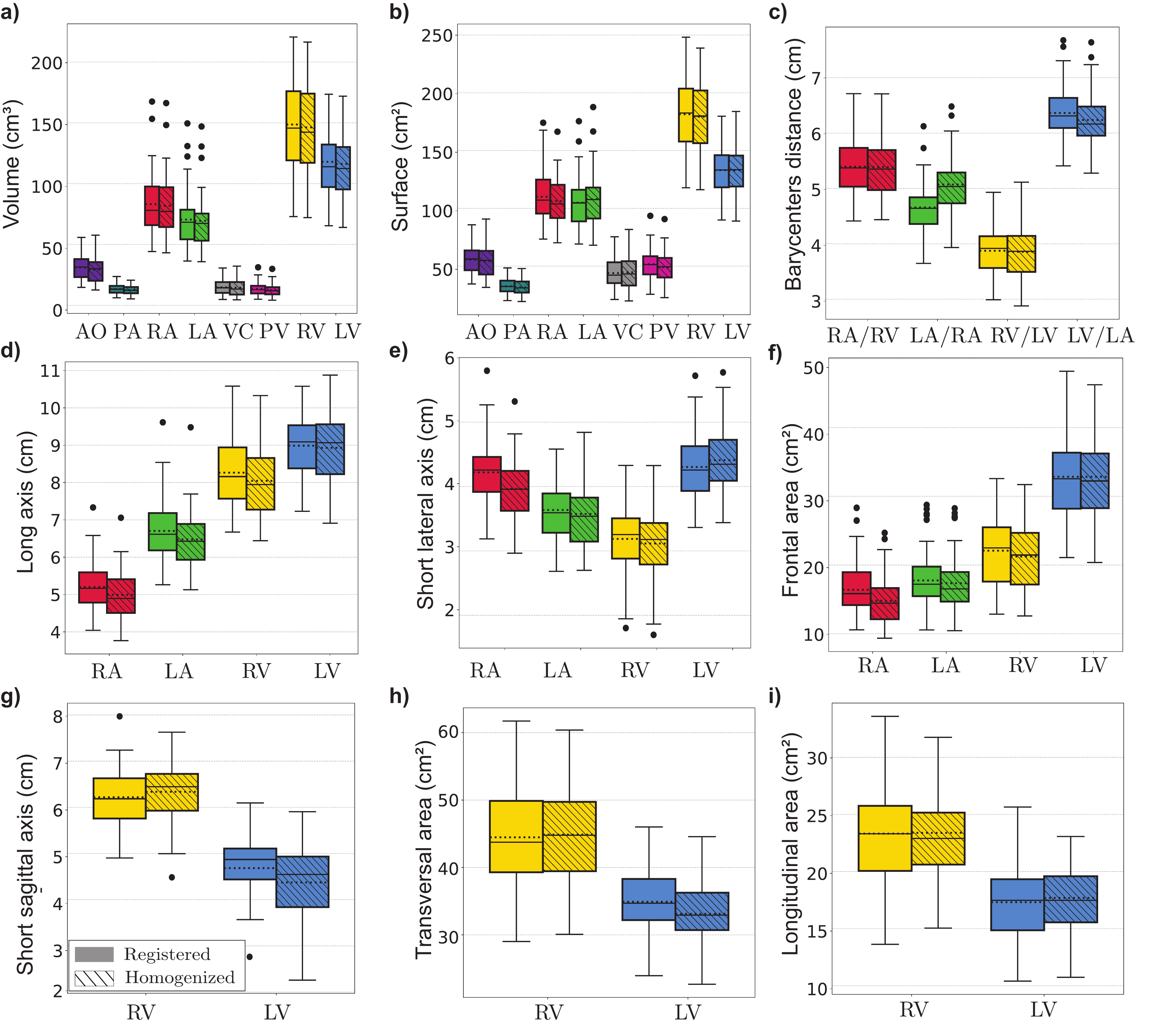}
    \caption{Box plots of anatomical and registered volumes (right) and surface (left) for all the anatomical landmarks. Central dotted lines represent means. A strong overlap between registered and anatomical distributions indicates a high degree of morphogeometric retention. }
   \label{fig:sur_vola}    
\end{figure}

A quantitative validation is performed to evaluate the extent to which the registration procedure preserves the patient-specific anatomical properties, using morphogeometric descriptors applied to the registered and homogenized datasets (see section~\ref{sec:evalm}).
The corresponding box plots for all the anatomical landmarks are shown in Figure~\ref{fig:sur_vola}, showing a good agreement between the distribution of each morphogeometric descriptors defined in section~\ref{sec:evalm}.
For each evaluation metrics, Table~\ref{tab:reconstruction_summary} reports the corresponding average $\mu_{\mathcal{R}}$ ($\mu_{\mathcal{H}}$) and standard deviation $\sigma_{\mathcal{R}}$ ($\sigma_{\mathcal{H}}$) of the registered (homogenized) dataset, along with the average samplewise absolute difference $\overline{|\Delta q|}$ and the reconstruction accuracy $R_a$.
\\ \indent
Overall, the average and standard deviation values match closely across the two datasets and reconstruction accuracy remains consistently high (always above 85\%, tipically around 95\%) across all structures.
The vessel surfaces have a $R_a=91-96\%$, with the lowest values in the PV and the highest in the PA, whereas higher reconstruction accuracy (exceeding $95\%$) is observed for the heart chambers (with one of the ventricles approaching $99\%$. A similar trend is observed for volumes,  with reconstruction accuracy within $87-93\%$ for vessels and above $97\%$ for heart chambers. The barycenters distance is also well preserved with $R_a\approx 91\%$ for the LA/RA and LV/LA and $R_a\approx 97\%$ for RV/LV and RA/RV, thus confirming a robust maintenance of ventricular alignment and overall inter-chamber configuration.
%\\ \indent
For the long axis, retention remains high across chambers, spanning roughly from a $\approx 93\%$ range in the RA to about $98\%$ in the LV, indicating strong preservation of longitudinal dimensions. The short lateral axis exhibits slightly greater variability, with the RA falling below $90\%$, while the other chambers remain at or above $95\%$, thus implying an overall strong preservation of the frontal area. 
For the short sagittal axis, RV has $R_a>92\%$, while the LV decreases to $R_a\approx 85\%$, indicating reduced agreement in short-axis measurements for the latter. Longitudinal and transversal areas further support the overall preservation of ventricular geometry.

%\begin{table*}[!htb]
\begin{sidewaystable}
\caption{Summary statistics comparing registered ($\mathcal R$) and homogeneous ($\mathcal H$) datasets for all morphogeometric descriptors $D$. For each anatomical region and dataset, average $\mu$ ($\pm$ standard errors SE) and standard deviations $\sigma$ are shown. Additionally, the average samplewise metric absolute difference $\overline{|\Delta q|}$ and the average reconstruction accuracy $R_a$ are reported.} \label{tab:reconstruction_summary}
\begin{tabular}{lc
S[ table-format=3.2(1.2),separate-uncertainty=true]
S[table-format=2.1]
S[ table-format=3.2(1.2),separate-uncertainty=true]
S[table-format=2.1]
S[table-format=-1.2]
S[table-format=2.1]
}
\toprule
{Descriptor $D$} & {Region} & {$\mu_\mathcal R \pm SE_\mathcal R$} & {$\sigma_\mathcal R$} & {$\mu_{\mathcal{H}} \pm SE_{\mathcal{H}}$} & {$\sigma_{\mathcal{H}}$} & {$\overline{|\Delta q|}$} & {$R_a$ [\%]} \\
\midrule
\multirow{8}{2cm}{Surface [cm$^2$]} & {\small LV} & 133.0(2.6) & 20.0 & 132.9(2.6) & 20.2 & 1.1 & 98.8 \\
{}                 & {\small RV} & 181.4(3.9) & 29.7 & 179.3(3.8) & 29.1 & 2.1 & 98.4 \\
{}                 & {\small LA} & 104.9(2.8) & 21.0 & 108.0(3.0) & 22.8 & 3.8 & 95.4 \\
{}                 & {\small RA} & 110.0(2.9) & 22.0 & 106.3(2.8) & 21.6 & 3.7 & 96.4 \\
{}                 & {\small AO} & 55.6(1.6)  & 12.0 & 54.4(1.8)  & 13.4 & 1.9 & 95.4 \\
{}                 & {\small PA} & 32.5(0.8)  & 5.9  & 31.6(0.8)  & 6.1  & 1.1 & 96.2 \\
{}                 & {\small PV} & 51.6(1.6)  & 12.3 & 49.0(1.6)  & 12.5 & 3.5 & 91.0 \\
{}                 & {\small VC} & 43.9(1.6)  & 12.3 & 44.7(1.9)  & 14.2 & 2.1 & 94.3 \\
\midrule
\multirow{8}{2cm}{Volume [cm$^3$]} & {\small LV} & 118.0(3.5) & 26.8 & 116.2(3.5) & 26.4 & 1.7 & 98.2 \\
{}                & {\small RV} & 148.7(4.8) & 36.2 & 146.3(4.6) & 35.3 & 2.4 & 98.2 \\
{}                & {\small LA} & 70.5(3.0)  & 23.3 & 69.5(3.1)  & 23.3 & 1.3 & 97.9 \\
{}                & {\small RA} & 83.2(3.3)  & 25.4 & 81.8(3.3)  & 25.0 & 1.4 & 97.8 \\
{}                & {\small AO} & 31.1(1.3)  & 9.8  & 29.2(1.3)  & 9.9  & 1.9 & 92.7 \\
{}                & {\small PA} & 13.3(0.5)  & 3.6  & 12.4(0.5)  & 3.5  & 0.8 & 92.8 \\
{}                & {\small PV} & 13.5(0.6)  & 4.9  & 12.5(0.6)  & 4.8  & 1.4 & 87.0 \\
{}                & {\small VC} & 14.8(0.8)  & 6.1  & 14.5(0.9)  & 6.7  & 0.7 & 93.3 \\
\midrule
\multirow{4}{2cm}{Barycenters distance [cm]} & {\footnotesize LV/LA} & 6.36(0.07) & 0.50 & 6.24(0.06) & 0.49 & 0.16 & 96.9 \\
{}                          & {\small RV/LV} & 3.87(0.06) & 0.43 & 3.86(0.06) & 0.46 & 0.06 & 98.2 \\
{}                          & {\small LA/RA} & 4.66(0.06) & 0.48 & 5.07(0.07) & 0.52 & 0.42 & 91.3 \\
{}                          & {\small RA/RV} & 5.39(0.08) & 0.58 & 5.39(0.07) & 0.55 & 0.10 & 97.7 \\
\midrule
\multirow{4}{1.8cm}{Long axis [cm]} & {\small LV} & 8.99(0.11) & 0.82 & 8.93(0.11) & 0.86 & 0.12 & 98.2 \\
{}                & {\small RV} & 8.26(0.12) & 0.91 & 8.05(0.13) & 0.96 & 0.24 & 95.9 \\
{}                & {\small LA} & 6.71(0.11) & 0.81 & 6.48(0.11) & 0.80 & 0.23 & 96.1 \\
{}                & {\small RA} & 5.20(0.08) & 0.63 & 4.99(0.09) & 0.65 & 0.25 & 93.5 \\
\midrule
\multirow{4}{2cm}{Short lateral axis [cm]} & {\small LV} & 4.30(0.07) & 0.52 & 4.41(0.07) & 0.52 & 0.12 & 96.2 \\
{}                 & {\small RV} & 3.19(0.07) & 0.55 & 3.12(0.07) & 0.55 & 0.10 & 96.2 \\
{}                 & {\small LA} & 3.64(0.07) & 0.53 & 3.57(0.07) & 0.52 & 0.12 & 95.4 \\
{}                 & {\small RA} & 4.22(0.06) & 0.49 & 3.96(0.06) & 0.47 & 0.35 & 89.1 \\
\midrule
\multirow{2}{2.2cm}{Short sagittal axis [cm]} & {\small LV} & 4.68(0.09) & 0.72 & 4.37(0.10) & 0.77 & 0.38 & 85.4 \\
{}                & {\small RV} & 6.22(0.08) & 0.64 & 6.34(0.10) & 0.78 & 0.32 & 92.5 \\
\midrule
\multirow{4}{1.7cm}{Frontal area [cm$^2$]} & {\small LV} & 32.9(0.8) & 6.1 & 32.8(0.8) & 6.1 & 0.4 & 98.5 \\
{}                      & {\small RV} & 22.1(0.7) & 4.9 & 21.2(0.6) & 4.7 & 0.9 & 95.0 \\
{}                      & {\small LA} & 17.8(0.6) & 4.2 & 17.4(0.6) & 4.3 & 0.5 & 96.6 \\
{}                      & {\small RA} & 16.4(0.5) & 3.7 & 14.8(0.5) & 3.4 & 1.6 & 87.3 \\
\midrule
\multirow{2}{2cm}{Longitudinal area [cm$^2$]} & {\small LV} & 34.9(0.7) & 5.5 & 33.1(0.7) & 5.2 & 2.1 & 91.1 \\
{}                           & {\small RV} & 44.5(1.0) & 7.6 & 44.9(1.1) & 8.1 & 2.4 & 93.2 \\
\midrule
\multirow{2}{2cm}{Transversal area [cm$^2$]} & {\small LV} & 17.3(0.4) & 3.0 & 17.7(0.4) & 3.1 & 0.5 & 95.7 \\
{}                          & {\small RV} & 23.3(0.6) & 4.7 & 23.4(0.7) & 4.9 & 0.8 & 95.3 \\
\bottomrule
\end{tabular}
%\end{table*}
\end{sidewaystable}

\subsection{Statistical shape modeling}\label{modelred}
\begin{figure}[htb!]
    \centering
    \includegraphics[width=0.95\linewidth]{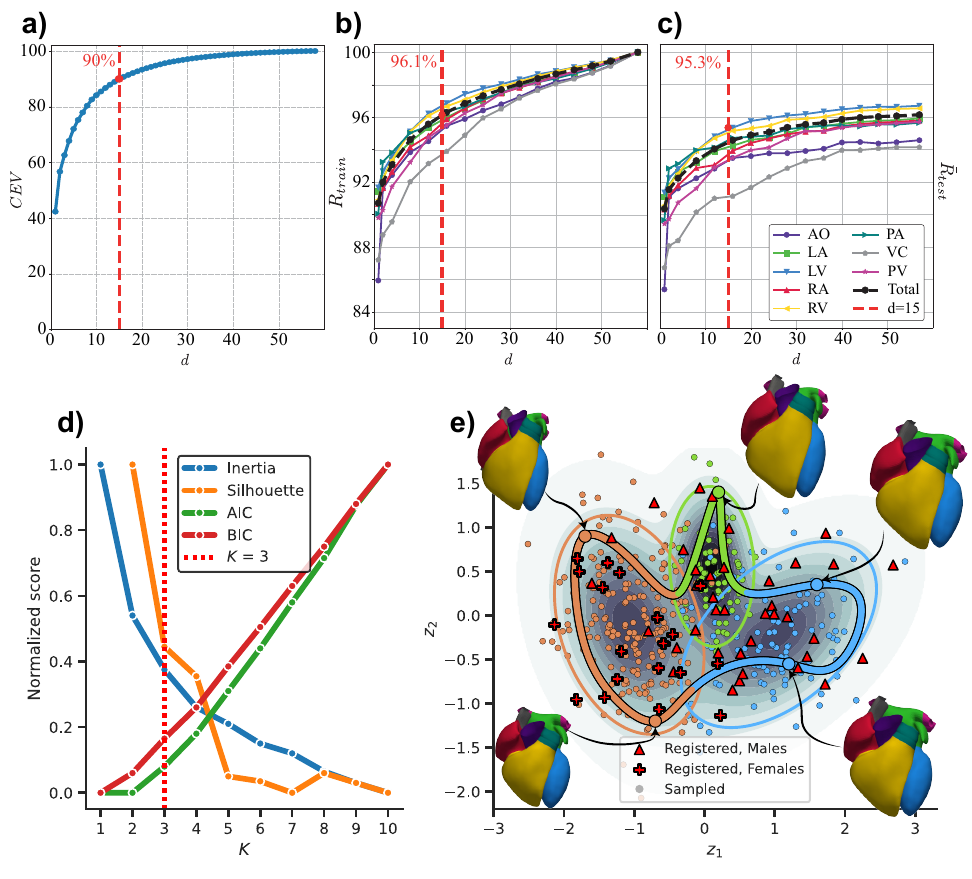}
    \caption{\textbf{Panel (a)}: Cumulative variance explained by each mode of the Principal component analysis. \textbf{Panel (b,c)}: Reconstruction accuracy $R_a$ with respect to the number of PCA modes ($d$) on the whole training set (Panel a) and test mesh in a leave-one-out (LOO) setting (Panel c). The colored lines indicate the reconstruction percentages for different anatomical regions. The vertical dotted red lines highlight the chosen number of modes $d=15$. \textbf{Panel (d)}: Cluster selection criteria normalized between $1-0$ with respect to the number of clusters. \textbf{Panel (e)}: Two-dimensional projection of the learned latent space, showing the Gaussian mixture density (background heatmap) and both registered and (GMM-) sampled instances (markers of the scatterplot). An example interpolation path is also shown, along with reconstructed heart meshes corresponding to sampled points along it.}
   \label{fig:pca_recon_train_test}    
   % : Orange $ \{(-0.7,-1.2), (-1.7,0.9)\} $, Green $ \{(0.2,1.4)\} $, and Blue $ \{(1.6,0.35), (1.2,-0.55)\} $
\end{figure}

A PCA representation of the registered dataset is now constructed.
% In order to evaluate the SSM, two key metrics are analyzed: compactness which is defined as the percentage of reconstruction accuracy (\ref{eq:rec_pca}) or cumulative variance
% explained (\ref{eq:cev}) by the model up to a certain number of modes and generalization, which evaluates the ability of the learned representation to perform on instances outside the training set and is expressed again as as the percentage of reconstruction accuracy.
Figure~\ref{fig:pca_recon_train_test}(a) shows the CEV (see section~\ref{sec:PCA}) computed on the whole registered dataset, with variance cumulatively explained by increasing the number of components. The characteristic elbow curve starts with a CEV of $42.3\%$ for $d=1$ and reaches a CEV of $90\%$ for $d = 15$.
\\ \indent
Figure~\ref{fig:pca_recon_train_test}(b,c) reports the reconstruction accuracy of the PCA-based shape model as a function of the number of modes ($d$) on the training set ${R}^{train}$, where the reconstruction is done on the same meshes used to train the PCA, and on the testing set $\bar{R}^{test}$ obtained under the LOO protocol. In both cases, the reconstruction accuracy is expressed separately for each cardiac chamber or vessel and for the global model.
% The reconstruction curves for the training subset show a rapid increase in accuracy within the first few components, highlighting the compactness of the model: a small fraction of modes captures the majority of anatomical variability within the subjects used to build the basis, reaching $96.1\%$ at $d=15$. The residual improvements from additional modes correspond to progressively finer or more localized additions.
Both training and test reconstructions start at approximately $90.4\%$ accuracy at $d = 1$, then increase rapidly before gradually leveling off. At $d=1$, the highest reconstruction accuracy is observed for the ventricular structures, indicating that the leading PCA mode primarily captures ventricular variability, which dominates the total dataset variance. The pulmonary artery follows, likely because only a short proximal segment is included in the SSM, which limits its geometric variability. The atria show slightly lower but still high reconstruction accuracy, whereas the aorta, pulmonary veins, and vena cava exhibit the lowest values. This anatomical ordering is largely preserved between the training and LOO test curves, suggesting consistent generalization across structures. The reconstruction percentages are substantially higher than the corresponding CEV values because CEV is computed after subtracting the mean geometry, whereas reconstruction accuracy is evaluated on the full geometry. Thus, the high reconstruction values indicate that the average registered anatomy already provides a strong geometric baseline for the parametric description, while the PCA modes account for the residual inter-subject variability around this mean shape.

While training reconstruction continues to improve as additional modes are included—consistent with the expected behavior of PCA when all components are retained—test performance plateaus at approximately  $96.4\%$, reaching $95.3\%$ already at $d=15$. This trend arises from the LOO evaluation protocol: the PCA basis is computed from a reduced training set, and excluding each test subject perturbs the covariance structure while removing deformation directions specific to that instance. As a result, test reconstruction saturates earlier, whereas training reconstruction continues to improve as model dimensionality increases.
%\\ \indent
The leading PCA components are primarily driven by anatomical regions that contribute the greatest share of variance, i.e., the ventricles, which control the largest portion of the mesh degrees of freedom. As a result, the global reconstruction curves largely mirror the ventricular ones.

\subsubsection{Gaussian Mixture Model on PCA }\label{GMM}
A population-level statistical representation of the anatomical variability
of the registered dataset is obtained through GMM over the PCA latent space.
The criteria defined in section~\ref{sec:model_selection} are computed as functions of the number of clusters $K$ and normalized by their maximum, as shown in Figure~\ref{fig:pca_recon_train_test}(d).
The inertia halves between $K=1$ and $K=2$, and decreases more slowly for higher $K$. Specifically, around $K=3$ the improvement in compactness begins to diminish, indicating that additional clusters provide only marginal gains.
On the other hand, the silhouette coefficient peaks at $K=2$ and it remains larger than $40\%$ at $K=3$, implying that cluster separation is still good.
\\ \indent
Moreover, the sharp rise of AIC and BIC for $K>3$ indicates that, although adding more clusters slightly improves the model fit, these gains are small compared ot the increase in model complexity.
Indeed, for a model with full covariance matrices, the number of parameters scales on the order of $K d^{2}$.
Therefore, given the limited size of the dataset and the PCA latent space set to $d=15$, $ K=3$ clusters are selected.

\subsubsection{Sampling}
\label{sec:sampling}
Figure~\ref{fig:pca_recon_train_test}(e) shows a two-dimensional projection of the GMM fitted in the reduced PCA space. The colored background represents the mixture probability density evaluated on the $(z_1,z_2)$ plane, obtained as the weighted sum of the projected Gaussian components. After fitting the GMM, 400 latent vectors were sampled from the learned distribution. The original registered anatomies are shown as red markers, with triangles denoting male subjects and crosses denoting female subjects. In contrast, sampled instances are shown as small points colored according to the Gaussian component from which they were generated.

The registered subjects exhibit a clear sex-related organization in the first two PCA coordinates. Male subjects are predominantly located toward positive $z_1$ values, whereas female subjects are mostly concentrated in the region of lower or negative $z_1$. This separation is consistent with the interpretation of the first PCA coordinate as a dominant size-related mode: positive $z_1$ values correspond to larger cardiac geometries, while negative values correspond to smaller anatomies. The overlap between male and female markers indicates that sex is not the only factor controlling the latent distribution; nevertheless, the scatterplot shows that global cardiac size contributes substantially to the stratification of the cohort in the leading PCA direction. In contrast, the spread along $z_2$ appears less directly associated with sex and more related to shape variability within each group.

Solid, color-coded ellipses represent iso-probability contours of the individual Gaussian components, defined by a constant Mahalanobis distance~\cite{asiatic1933journal}, here chosen equal to $4$ enclosing approximately $86\%$ of the probability mass of each component. The orientation and elongation of each ellipse are determined by the eigenvectors and eigenvalues of the projected covariance matrix, respectively. The resulting mixture identifies three main high-density regions in the $(z_1,z_2)$ plane: a left cluster associated with smaller geometries (corresponding mainly to the registered female subjects along $z_1$), a central cluster with intermediate size and larger spread along $z_2$, and a right cluster corresponding to larger geometries. The sampled points reproduce both the location and dispersion of the original registered instances, indicating that the GMM captures the main structure of the observed latent distribution while enabling the generation of additional anatomically plausible samples.

An interpolation trajectory is then defined in the full PCA space and projected onto the $(z_1,z_2)$ plane. Selected points along this trajectory, indicated by bigger points, are reconstructed using the PCA mapping in \eqref{eq:recPCA}. The corresponding cardiac meshes are displayed around the latent-space projection. The trajectory is chosen to traverse the peripheral regions of the Gaussian components, near the covariance-contour boundaries, to visualize the anatomical effects of structured variations along the dominant latent directions. The reconstructed geometries show that moving along the $z_2$ direction induces significant structural changes to the RV, which becomes progressively more elongated while the global scale remains largely retained. As the trajectory enters the green region, characterized by increasing $z_1$, the deformation pattern shifts toward a global size increase, with minimal additional modification of the RV morphology. Upon entering the blue region, characterized by persistently large $z_1$ values, the path simultaneously contracts along $z_2$, leading to a gradual reduction of the previously observed elongation. Correspondingly, the RV geometry progressively recovers a configuration more similar to the initial orange shapes, despite the maintained global enlargement.

These observations indicate that distinct latent directions encode qualitatively distinct, although not fully independent, anatomical effects: variations along $z_2$ predominantly modulate RV structural characteristics, whereas $z_1$ primarily governs global scaling. The observed reconstruction dynamics are consistent with the dataset's statistical structure, in which the ventricles account for most of the mesh degrees of freedom and exhibit the greatest variability. Consequently, global deformation trends closely follow ventricular behavior, particularly that of the RV, as discussed in section~\ref{modelred}, reflecting its comparatively larger surface, volume, and dispersion in the morphogeometric analyses.

\subsubsection{Statistical analysis}
\begin{figure}[htb!]
    \centering
    \includegraphics[width=0.99\linewidth]{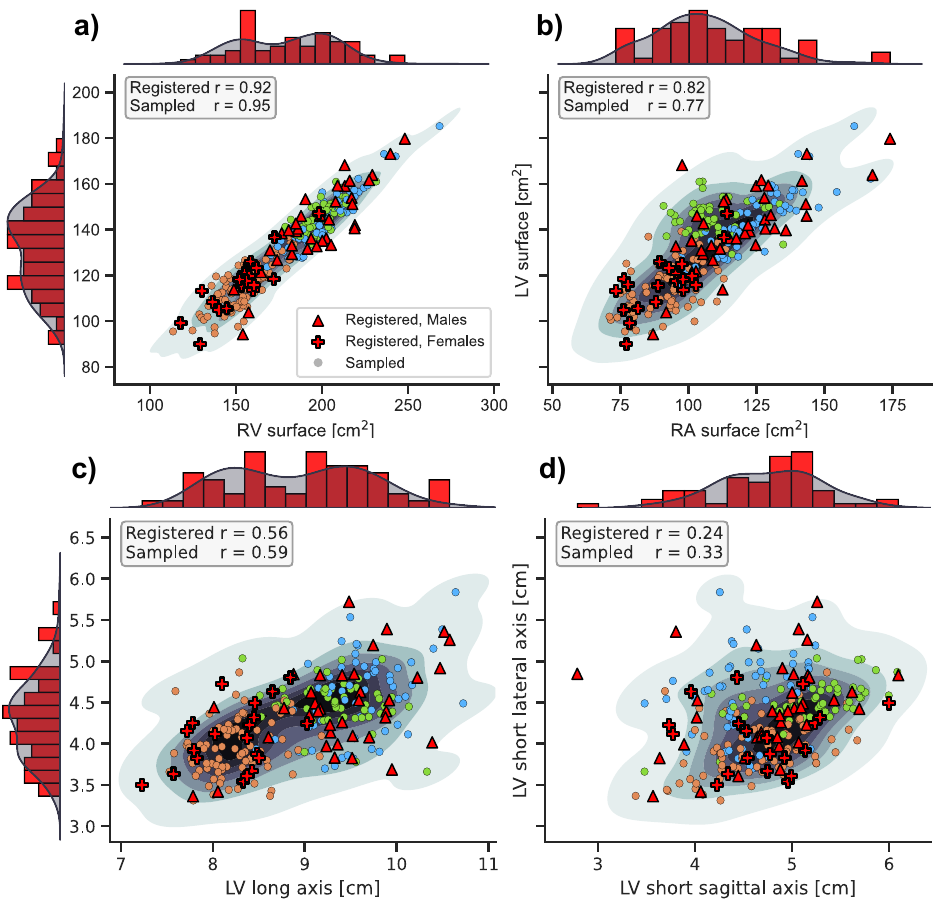}
    \caption{Joint probability distributions of different morphological descriptors evaluated sampling 200 shapes from the fitted GMM. Scatter plots of these samples (small points), as well as the original registered shapes (red triangles and crosses) are shown. Marginal distributions of both sampled (dark interpolated curve) and registered shapes (red histogram bars) are shown for each joint distribution. Panels a) and b) show the surface area of the right ventricle-left ventricle and right atrium-left ventricle. Panels c) and d) show the long-short lateral and short sagittal-short lateral axes lengths of the left ventricle. Strong correlations are evident for panels a) and b), while moderate/weak for panels c) and d).}
    \label{fig:joint}
\end{figure}
By generating additional samples that follow the learned latent distribution, the proposed framework effectively augments the original dataset, enabling the study of statistical trends in an expanded yet coherent shape space. Rather than analyzing isolated reconstructions, this augmented space allows a more robust examination of relationships between anatomical measurements using the metric defined in section~\ref{sec:evalm}.

As an example, Figure~\ref{fig:joint} compares the distribution of selected morphogeometric descriptors obtained from the registered original PCA registered instances (red triangles for males and red crosses for females) and with 200 randomly sampled instances drawn from the fitted GMM in latent space (small circles colored based on the Gaussian cluster). The dark isocontours show the kernel density estimated from the sampled instances.

Panels (a) and (b) show joint distributions of cardiac chamber surface areas for paired anatomical structures (LV, RV and RA). The registered subjects display a clear sex-related separation in these surface-based measurements, suggesting that gender is highly correlated with both ventricular and atrial chamber size. The density estimates indicate strong positive correlations between the LV and RV surfaces, and between the LV and RA surfaces, with Pearson correlation coefficients of approximately $R=0.95$ and $R=0.8$, respectively. The stronger ventricular correlation is visually reflected by the narrow, elongated density contours in panel (a), indicating that the RV surface area varies within a relatively restricted range for a given LV surface area. In contrast, the broader contours in panel (b) show that, for a fixed LV surface area, the RA surface area exhibits greater variability.

The marginal distributions reported along the axes of Figure~\ref{fig:joint} also show non-Gaussian, partially bimodal patterns for the chamber surface areas. This behavior is consistent with the multimodal structure imposed by the GMM fitted in the SSM latent space. Since the probabilistic model is defined on whole-heart shape representations, samples drawn from the GMM concentrate around multiple high-density regions of the learned anatomical distribution. When surface measurements are subsequently computed from these sampled geometries, the multimodal latent structure is reflected in the derived morphogeometric descriptors. The agreement between sampled and registered instances (both in terms of distribution shape and Pearson correlation coefficient) further indicates that the GMM preserves the main correlation structure observed in the registered cohort.

Panels (c) and (d) show the joint distributions of the LV principal axes. The scatterplots of the registered PCA instances highlight that a clear separation between different genders is provided by the LV long axis. Instead, short lateral and sagittal axes display an almost complete overlap between sexes. This indicates that male and female LV geometries differ mainly in long-axis length, while transverse dimensions show greater inter-subject overlap and weaker discriminatory capacity. The gender-agnostic distributions show that the LV long and short lateral axes are moderately correlated ($R=0.59$), indicating that larger LV long-axis dimensions are generally associated with larger short lateral-axis dimensions. By contrast, the LV short sagittal and lateral axes show weak correlation, with $R=0.33$, as also suggested by the more compact and nearly circular density contours in panel (d). Overall, the LV shapes in the cohort span long-axis lengths of approximately $7$--$10.5$ cm, whereas the short lateral and sagittal axes vary over narrower, more weakly coupled ranges of approximately $3.5$--$6$ cm. This indicates that the dominant LV shape variability occurs along the long axis, while the transverse dimensions can vary more independently, giving rise to both more elongated and more circular cross-sectional configurations.

\subsubsection{Sensitivity analysis}\label{sec:results_sens_analysis}
%\begin{figure}[htb!]
%  \centering
%    \includegraphics[width=\textwidth]{figure10.pdf}
%    \caption{Sensitivity analysis of morphogeometric descriptors with respect to PCA modes $z_i$. Each sensitivity value $S_i$ is computed as the absolute value of the first-order ANCOVA sensitivity index (see section~\ref{sec:sens_analysis}), expressed as a percentage (0--100\%), and rounded to the nearest integer.}
%   \label{fig:pca_modes_sensitivity}   
%\end{figure}

Extended Data Figure~\ref{fig:pca_modes_sensitivity} depicts the heatmap of the ANCOVA sensitivity indices described in subsection~\ref{sec:sens_analysis} for each chamber volume, inter-barycenter distances and the principal axes lengths: long, short lateral and short sagittal. As expected, each PCA mode simultaneously influences multiple morphogeometric descriptors. This behavior reflects the nature of PCA, where modes are constructed to hierarchically maximize explained variance rather than to isolate distinct anatomical features. Consequently, individual modes encode coupled geometric variations affecting several descriptors simultaneously.

The first PCA mode accounts for most of the sensitivity of scale-related descriptors. This is particularly evident for chamber volumes, with the exception of the pulmonary veins and vena cava, which appear less directly coupled to global heart size. The same mode also explains most of the variability in the inter-chamber barycenter distances and contributes substantially to the long-axis descriptors. These results indicate that the leading mode mainly represents a global size-like deformation of the whole-heart geometry.

In contrast, the chamber principal axes, especially the short lateral and sagittal axes, exhibit a more distributed sensitivity pattern across higher-order modes. This suggests that, although the first mode captures most of the variance associated with global volume changes, more localized and shape-specific deformations are encoded by subsequent PCA modes. The pulmonary veins and vena cava also show stronger sensitivity to higher-order modes, consistent with the reconstruction analysis in Figure~\ref{fig:pca_recon_train_test}, where these regions exhibited lower reconstruction accuracy than the main cardiac chambers. This likely reflects the fact that the dominant PCA directions are primarily governed by the larger ventricular and atrial structures, while secondary vessel variability is represented by modes of higher order.

The short lateral-axis descriptors are still partially influenced by the first mode, whereas the short sagittal-axis descriptors are almost entirely controlled by higher-order modes and display a sparse sensitivity distribution.

\section{Discussion}
This work presents a semi-automatic pipeline for generating simulation-ready cardiac meshes directly from CT data. By combining a template-based, multi-chamber registration strategy with a multi-scale optimization scheme, the framework yields watertight, isotopological meshes that preserve subject-specific anatomical variability while ensuring numerical robustness and structural consistency. 

The full cohort of 58 anatomies was generated in approximately 15 hours using only one GPU, including the iterative template-update procedure. The processing time could be further decreased by using one GPU per patient, which would enable the processing of the full cohort in approximately 16 minutes. This represents a substantial reduction in manual processing time, since producing a comparable set of simulation-ready cardiac meshes through manual or case-by-case mesh repair would require several weeks of expert work and would introduce additional operator-dependent variability.

The code supporting this work has been made publicly available as open-source software at \url{gitlab.com/gssi-fluids/heart-ssm}. The released framework enables the reconstruction, regularization, and parametrization of new patient-specific image data, producing high-quality cardiac meshes suitable for downstream in silico studies in a few minutes. In addition to the processing pipeline, the statistical shape model developed in this study, consisting of the PCA basis and the fitted GMM in latent space, is released with the code. This provides an accessible resource for generating clean and anatomically variable heart geometries without requiring users to repeat the full cohort-construction procedure. 

The set of morphogeometric descriptors quantitatively demonstrates strong preservation of both global and chamber-level anatomical characteristics. Indeed, the anatomical validation study suggests that the registration procedure preserves most geometric evaluation metrics, exceeding $85\%$ reconstruction accuracy. This suggests that the dominant chamber morphology is preserved, while smaller structures and local cross-sectional descriptors remain more sensitive to anatomical variability, segmentation uncertainty, and the smoothing inherent to template-based deformation. Indeed, the 85\% value should be interpreted as a conservative estimate of the actual registration fidelity, since the reference measurements are extracted from the raw segmentations, which may contain mesh defects, local discontinuities, small holes, or irregular chamber boundaries. Such imperfections can affect the computation of geometrical descriptors, thereby introducing apparent discrepancy between registered and anatomical geometries. Accordingly, the selected regularization strength $\gamma$ was chosen to reflect the accuracy of the segmented targets. In the presence of target meshes with higher anatomical accuracy and fewer segmentation artifacts, the regularization strength could be reduced, allowing the optimization to enforce a closer geometric match between the homogenized and registered instances. Conversely, for the present dataset, a stronger regularization was preferable to avoid overfitting local segmentation defects and to preserve smooth, anatomically coherent, and simulation-ready geometries.

The enforced point-to-point correspondence across reconstructed geometries enables the construction of a statistical shape model of the whole heart, along with the main cardiac arteries/vessels. PCA reveals that a compact latent space of size $d=15$ captures the dataset's population variability. In the present cohort, the leading latent mode appears to encode a global size-related deformation, affecting chamber volumes, surface areas, inter-chamber distances and ventricles and atria long axes. Contrarily, higher-order modes contribute to more localized variations in vessel geometry and chamber shape, including changes in secondary chamber axes (short lateral and sagittal) and cross-sectional morphology. Considering the morphological descriptor sensitivity analysis conducted for each latent parameter, the majority of this global scale-related geometrical variability is associated solely with the first mode. Therefore, when the downstream analysis is mainly affected by scale-related anatomical differences, a very low-dimensional parametrization may be sufficient to describe the cohort. Conversely, studies sensitive to finer geometric features require additional PCA modes to capture the relevant variability (at least the first five modes in the present study). Probabilistic modelling via the Gaussian mixture model detects data clustering and statistical trends, such as the strong linear relationship between RV and LV surface measurements, with a Pearson correlation coefficient of approximately $0.95$.

Although these statistical results are expected to depend on the specific CT dataset used (which includes a relatively limited number of anatomies, $m=58$), the proposed method is general and can be straightforwardly extended to larger datasets. 

Indeed, a natural continuation of this work is the application of the full pipeline (fast segmentation, iterative registration, PCA and statistical analysis) to a larger patient-based dataset. A broader cohort would provide a more representative anatomical basis for template refinement and statistical modeling, reducing the influence of individual outliers and improving the robustness of the learned latent space structure, Gaussian mixture components, and, lastly, the shape variability. In this sense, the proposed framework should be regarded as scalable and progressively improvable: as additional medical images are incorporated, the registration template, PCA basis, and shape model can be recalibrated to better capture the anatomical variability of the target population. Consequently, the recalibrated PCA will allow for the generation of on-demand virtual cohorts of cardiac anatomies reproducing the statistical variability of the population at large.

In this context, the statistical model supports the generation of anatomically realistic virtual cohorts, facilitating in-silico studies, uncertainty quantification, and the investigation of population-level trends in cardiac function and pathology. Indeed, the availability of consistent, simulation-ready meshes with point-wise correspondence opens the door to integrating electrophysiological and hemodynamic simulations within a unified population-based framework.

Despite the demonstrated flexibility of the proposed framework, registration of novel cardiac anatomies from large datasets should be performed under supervision as template-based deformation may introduce an implicit bias toward the reference anatomy and limit the representation of rare or pathological morphologies. Nevertheless, template-based approaches can still be extended to more complex anatomical scenarios by using a library of reference templates rather than a single baseline geometry. In the presence of extreme cardiac shapes or topology-altering pathologies, the most appropriate template could be selected before registration according to its geometric and topological similarity to the target anatomy. This strategy would preserve the advantages of template deformation, including mesh quality and point correspondence, while broadening the range of anatomies that can be accurately represented.

The methodology has been applied to the reconstruction of the endocardial surfaces of the cardiac chambers, as the contrast resolution of standard CT imaging does not reliably allow accurate estimation of myocardial wall thickness or vessel wall structures. Consequently, the proposed procedure could be extended to more advanced imaging datasets where such information is available. Alternatively, wall thickness could be incorporated a posteriori by offsetting the reconstructed endocardial surfaces using locally estimated thickness fields or prior knowledge derived from anatomical atlases, enabling the generation of more physiologically complete geometries.

Another promising research direction concerns the mathematical representation of cardiac shapes. In the present formulation, registration is performed on explicit surface meshes by deforming a fixed template. This strategy enforces common topology and point-to-point correspondence across the cohort, which is essential for PCA-based shape modelling and directly supports the generation of simulation-ready meshes. However, it also restricts the admissible anatomical variability to geometries that can be represented as smooth deformations of the chosen template. This limitation may become relevant when extending the framework to more heterogeneous populations or to pathological anatomies characterized by substantial topological or structural differences. More flexible implicit representations, such as neural signed-distance functions, as used in~\cite{KONG2024103293} for congenital heart defects, or related continuous shape embeddings, could relax this constraint and provide a more general representation of anatomical variability. Nevertheless, their integration into simulation-oriented pipelines remains non-trivial. In particular, the template in the present framework not only provides geometry, but also carries correspondence, anatomical labeling, and mesh-quality information. Therefore, implicit or learned shape representations would still need to preserve, transfer, or reconstruct this information to remain compatible with statistical analysis, regional measurements, and downstream numerical simulations.

A related direction is the use of self-supervised machine-learning methods to establish point correspondence across shapes, such as~\cite{adams2026point2ssm++}. Learned correspondence models could replace hard template-based matching with a softer and more flexible criterion, potentially enabling partial registration of geometries with different topologies, missing structures, or local inconsistencies. However, these methods have mainly been demonstrated on anatomies with relatively simple or moderately complex topology, whereas whole-heart geometries involve multiple chambers, vessel openings and complex inter-structure interfaces. Their robustness in such anatomically complex, simulation-driven settings therefore remains to be validated.

Finally, more powerful unsupervised machine-learning approaches could be integrated for generative shape modelling, either complementing or replacing PCA. Nonlinear generative methods, including variational autoencoders~\cite{kingma2013auto}, normalizing flows~\cite{kobyzev2020normalizing}, generative adversarial networks~\cite{goodfellow2014generative}, and diffusion models~\cite{ho2020denoising}, may capture anatomical variability beyond a linear PCA basis, particularly in larger and more heterogeneous cohorts, as demonstrated in \cite{fabbri2026graph}. Nevertheless, these models require substantially larger curated datasets to avoid overfitting and ensure anatomically plausible generation. For this reason, PCA remains a robust, interpretable, and data-efficient baseline for medical SSMs, while nonlinear generative models represent a promising extension as larger simulation-ready datasets become available.

\section{Methods}\label{sec:Methods}
\begin{figure*}[!htb]
    \centering
    \includegraphics[width=0.99\textwidth]{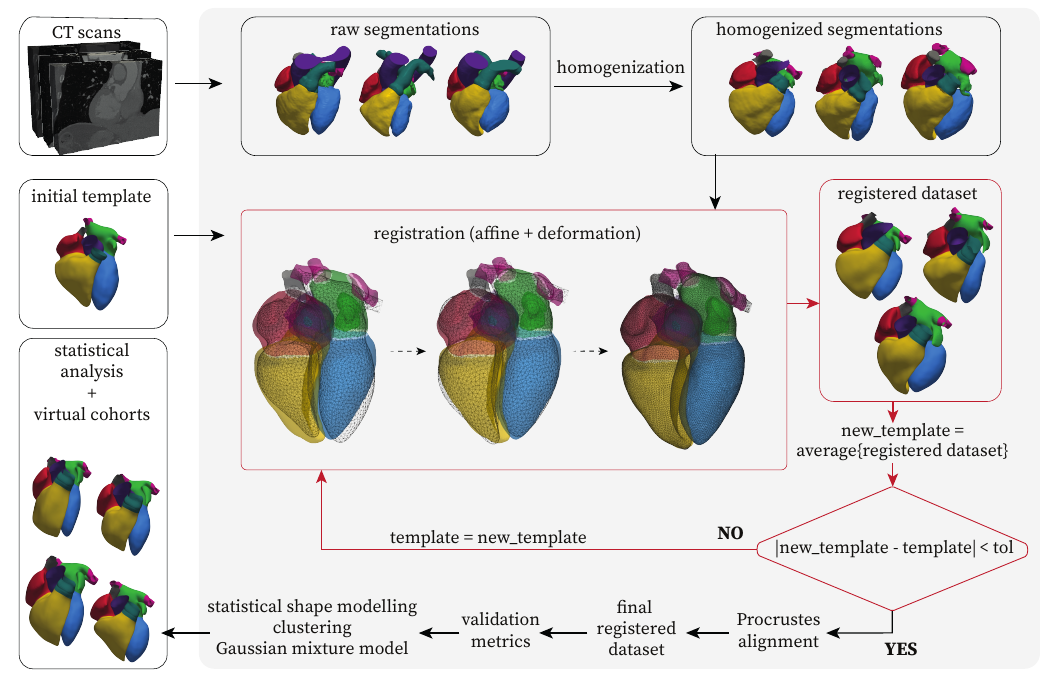}
    \caption{High-level workflow of the proposed pipeline for population-based cardiac modeling. \textbf{Fast segmentation}: CT scans are first processed to extract multi-chamber heart geometries from the images using AI-based tools. \textbf{Homogenization}: A second step involves a series of preprocessing operations, e.g., trimming the anatomies to the objective region of interest. \textbf{Registration}: Shapes are standardized via affine transformations; the template is non-rigidly deformed by minimizing a Chamfer-distance functional to match each anatomy. \textbf{Average and Update template}: The template is iteratively refined by comparison with the population average until convergence. \textbf{Statistical shape modelling}: The final set of topologically consistent, simulation-ready meshes enables shape modelling. Generalized Procrustes Analysis (GPA) to remove residual pose variability and ensure alignment across the population. A reduced representation is built using PCA and clustering, which supports the construction of a Gaussian Mixture Model over the PCA space. Sampling from this model enables the creation of synthetic cohorts and supports quantitative statistical analyses.}
    \label{fig:big}
\end{figure*}

% 1 segmentazione problematica, template objective
This section describes the workflow used to generate anatomically consistent cardiac meshes and a population-level statistical shape model from a dataset of CT scans. The whole pipeline is shown in Figure~\ref{fig:big}. Section~\ref{sec:dataset} describes the preliminary segmentation process applied to a clinical CT dataset. The registration of a template cardiac mesh on the segmented anatomies is presented in section~\ref{sec:registration}, while section~\ref{sec:evalm} defines the morphogeometric descriptors used for a quantitative validation of the registration step. Finally, section~\ref{sec:ssm} details the statistical shape modeling approach based on PCA and GMMs for dimensionality reduction, clustering, and sampling.

\subsection{Fast segmentation and homogenization}\label{sec:dataset}

We analyzed a dataset of 58 CT scans acquired at the European Hospital of Rome from patients with suspected coronary stenosis who subsequently underwent coronary angiography with negative findings. The resulting cohort therefore represents subjects with no evidence of significant coronary artery disease and can be considered as a reference population of anatomically healthy hearts.

All images were obtained using electrocardiogram (ECG)-gated CT acquisition protocols. For consistency, only image phases corresponding to approximately 80\% of the cardiac cycle were retained, representing the end-diastolic phase. Due to acquisition variability, the exact phase ranged between 70\% and 80\% across subjects, with a mean value of $78 \pm 3$\%. In the following, quantitative values are reported as mean $\pm$ standard deviation inside parentheses. The age of the subjects ranged from 40 to 79 years ($61 \pm 8$ years). The cohort included 38 male and 20 female subjects. All CT volumes had a fixed in-plane dimension of $512 \times 512$ pixels, while the number of slices varied across acquisitions ($249\pm46$), depending on the anatomical region covered. The scanned volume encompassed a region centered on the coronary arteries, often including the entire heart and proximal segments of the major vessels. The field of view had a dimension of approximately $242 \times 242 \times 153$ mm$^3$. The voxel spacing is constant across subjects at $0.47 \times 0.47 \times 0.62$ mm$^{3}$/voxel. Due to the finite acquisition field of view, elongated anatomical structures were occasionally truncated. This results in variably cut vessels across subjects, most notably affecting the aortic arch and the descending aorta, and introduces additional variability in the extent of the segmented anatomy.

A preliminary segmentation is first performed on each of the CT scans, using the TotalSegmentator tool \cite{Wasserthal2023TotalSegmentator} through the 3DSlicer software \cite{FEDOROV20121323}. Specifically, the Heart Chambers HighRes module was used to obtain high resolution segmentation at $0.5$ mm per voxel in each direction. This tool enables the automatic detection and segmentation of each cardiac chamber together with the aorta and pulmonary artery. The resulting segmentation of the cardiac anatomy is then refined using semi-automatic 3DSlicer tools: the pulmonary veins, left atrial appendage and superior/inferior vena cava are manually segmented using intensity-based thresholding and pruning. Vena cava and pulmonary veins are trimmed at a standard length of approximately 10~mm. Following the segmentation phase, these anatomies are converted into surface geometries using the Flying Edges isosurface extraction algorithm \cite{schroeder2015flying}, producing triangulated meshes for each anatomical structure. An example of the resulting raw, segmented mesh is shown in Figure~\ref{fig:defects}, highlighting typical geometrical imperfections after this semi-automatic segmentation phase.

As anticipated in this section, owing to differences in the sizes of the scanning windows, most anatomies do not contain the entire aortic arch, and the level of truncation varies from case to case. Therefore, the raw segmented meshes are homogenized to have roughly the same anatomical domain. In particular, the aorta and pulmonary artery are trimmed through an automatic procedure retaining always the same relative length. The trimming criterion was defined based on the local vessel diameter $l$: segments exceeding $0.7\,l$ for the pulmonary artery and $1.5\,l$ for the aorta were removed.

To eliminate variability due to rigid transformations, all geometries were subsequently mapped into a common reference frame. The origin of this reference system was defined as the centroid of the left ventricle (LV). The $x$-axis was oriented along the vector connecting the right ventricle (RV) centroid to the LV centroid. Finally, to resolve the remaining rotational degree of freedom, a rotation about the $x$-axis was applied such that the vector connecting the LV and left atrium (LA) centroids was aligned with the $y$-axis. The final result of these transformations is a set of 58 cardiac geometries which will be referred to as ``homogenized dataset".

Overall, the time required to process a patient CT image and obtain the corresponding homogenized surface mesh is mainly governed by the first two stages of the pipeline: AI-based automatic segmentation and semi-automatic refinement in 3D Slicer to include the proximal portions of the peripheral vessels. The automatic segmentation step requires only a few minutes on a moderately performing laptop, whereas the subsequent semi-automatic refinement can require up to approximately 15 minutes per case. The remaining preprocessing operations, including remeshing and automatic clipping, are fully automatic and require only a few seconds.

\subsection{Registration}\label{sec:registration}
Although the set of raw segmentations detailed in the previous subsection characterizes the morphological variability of the heart, they are not suitable for multiphysics simulations or statistical shape modeling, owing to the presence of several geometrical defects. This issue is addressed by designing a registration procedure that deforms a template mesh to match the segmented anatomies. The approach simultaneously preserves the geometric regularity of the template mesh while capturing the patient-specific cardiac anatomy.
\\ \indent
At the initial stage of the procedure, the template $\mathcal{M}_{\mathcal{T}}$ is a single simulation-ready, high-quality mesh of the anatomical region of interest. The latter is obtained from a CT scan of a healthy subject from a previous study and subsequently refined through an extensive manual process aimed at removing the segmentation and meshing defects described above. This curation steps required at least one week of work by a highly trained operator. The mesh is represented as an undirected graph $\mathcal{M}_{\mathcal{T}} = (V_{\mathcal{T}}, E_{\mathcal{T}}, A_{\mathcal{T}})$, where  $V_{\mathcal{T}}$ is the set of its  $N_{\mathcal{T}}$ vertices, having spatial coordinates $\mathbf{x}_{\mathcal{T}}$ (each one lying in $\mathbb{R}^3$). The vertex connectivity is given by the set of edges $E_{\mathcal{T}}$ and adjacency matrix $A_{\mathcal{T}}$.
\\ \indent
On the other hand, each homogenized cardiac anatomy retrieved from the CT dataset, is also represented as a triangulated mesh  $\mathcal{M}^i_{\mathcal{H}} = (V^i_{\mathcal{H}}, E^i_{\mathcal{H}}, A^i_{\mathcal{H}})$, where $V^i_{\mathcal{H}}$ is the set of its $N^i_{\mathcal{H}}$ vertices of the $i$-th segmentation, having spatial coordinates $\mathbf{x}^i_{\mathcal{H}}$ (each one lying in $\mathbb{R}^3$).
Hence, different segmentations generally exhibit distinct numbers of vertices $N^i_{\mathcal{H}}$ and different connectivity $(E^i_{\mathcal{H}},~A^i_{\mathcal{H}})$.
\\ \indent %nota ho scritto tutta la deformazione (anche affine sul template)
Firstly, for each homogenized segmentation, the template undergoes an affine transformation (with scale factor $\mathbf{s}_i$ and rotation matrix $\mathbf{R}_i$) to minimize the distance between its chamber barycenters with respect to the ones of the segmentation,  using the Kabsch-Umeyama algorithm \cite{kabsch1976solution}.
This results in a rotated and stretched template mesh $\mathcal{M}^i_\mathcal{T}$ maintaining the same connectivity but with node coordinates modified by the affine transformation into $\mathbf x^i_{\mathcal T}$.
Then, a deformation step is performed to morph the template over the homogenized segmentation. The deformation is formulated as a minimization problem of a functional $\mathcal{D}^i$ summing the distance among the heart chambers of $\mathcal{M}^i_{\mathcal{T}}$ and $\mathcal{M}^i_{\mathcal{H}}$:
\begin{equation}
\min_{\mathbf{x}^{i}_{\mathcal{T}} \in V_{\mathcal{T}}} \mathcal{D}^i(\mathcal{M}^i_{\mathcal{T}}, \mathcal{M}^i_{\mathcal{H}})=\min_{\mathbf{x}^{i}_{\mathcal{T}} \in V_{\mathcal{T}}} \;
\sum_{c=1}^{n_c} \mathcal{D}^{i,c}\!\left(\mathcal{M}^i_{\mathcal{T}}, \mathcal{M}^{i}_{\mathcal{H}}\right),
\label{eq:min_prob}
\end{equation}
where $c$ identifies the considered anatomical region (or heart chamber), ranging here from $1$ to $n_c=8$ (four heart chambers plus four vessels).
Our choice of the distance functional is the Chamfer distance, defined as follows:
\begin{equation}
\label{eq:cd_formal}
\mathcal{D}^{i,c}(\mathcal{M}^i_{\mathcal{T}},\mathcal{M}^i_\mathcal{H})
=
\frac{1}{N^c_{\mathcal{T}}}
\sum_{\mathbf{x}^{i}_{\mathcal{T}}\in V^c_{\mathcal{T}}}
\min_{\mathbf{x}^i_{\mathcal H}\in V^c_i}
\|\mathbf{x}^{i}_{\mathcal{T}}-\mathbf{x}^i_{\mathcal{H}}\|_2^2
\;+\;
\frac{1}{N^c_i}
\sum_{\mathbf{x}^i_{\mathcal H}\in V^c_i}
\min_{\mathbf{x}^{i}_{\mathcal{T}}\in V^c_{\mathcal{T}}}
\|\mathbf{x}^{i}_{\mathcal{T}}-\mathbf{x}^i_{\mathcal{H}}\|_2^2,
\end{equation}
with
$\mathbf{x}^i_{\mathcal{T}}$ and $\mathbf{x}^i_{\mathcal{H}}$  the coordinates of the $N^c_{\mathcal{T}}$ template and $N^c_i$ homogenized  mesh nodes belonging to the chamber $c$.
The Chamfer distance (Equation~\eqref{eq:cd_formal}) evaluates the nearest-neighbor distance from each point in one set to the other set, thus promoting a reconstruction that follows the spatial structure of the objective geometry. Although the loss terms are defined over each cardiac chamber,  the minimization of Equation~\ref{eq:min_prob} is performed using a preconditioned gradient descent scheme \cite{10.1115/1.2746378}, where the preconditioning operator introduces a diffusion of the gradient across the mesh connectivity, thereby enforcing spatial coherence of the updates. The template vertex positions are updated according to:
$\mathbf{x}^{i}_\mathcal{T} \leftarrow \mathbf{x}^{i}_\mathcal{T} - \eta (I + \gamma L)^{-1} \partial \mathcal{D}^i / \partial \mathbf{x}^{i}_\mathcal{T}$, where $L$ denotes the combinatorial Laplacian of the template mesh and $(I + \gamma L)^{-1}$ acts as a preconditioning operator. A learning rate $\eta=0.01$ is used, following previous analysis in \cite{scarpolini_reg}. The preconditioning operator acts as a regularization constraint that balances between the minimization of the chamfer distance and the smoothness of the resulting deformation field. As a result, the vertex displacements are smoothed to accomodate realistic deformations, for example in the transitional regions between chambers. Consequently, the parameter $\gamma$ controls the strength of this smoothing mechanism, with larger values yielding a more globally regular deformation, while smaller values allow greater sensitivity to local geometric details of the target segmentation. The limiting case $\gamma = 0$ reduces the scheme to a standard gradient descent.

To capture geometric features at different spatial scales, the deformation approach in Equation~\ref{eq:min_prob} is applied sequentially at multiple mesh resolutions, following the strategy proposed in~\cite{scarpolini_reg}. In the first stage, a coarse template mesh with 4000 vertices is deformed until the Chamfer-distance functional in~\eqref{eq:cd_formal} reaches a plateau. The resulting displacement field is then interpolated onto the original, higher-resolution template mesh (the input one), and the optimization is restarted to refine local anatomical details while preserving the large-scale deformation already recovered at the coarse level (see Figure~\ref{fig:loss}). Following the heuristic considerations described above regarding the preconditioning operator, the smoothing parameter was set to $\gamma=120$ for the coarse stage and $\gamma=80$ for the finer resolution one.

It should be noted that the deformation field applied to the template mesh $\mathcal{M}_{\mathcal{T}}$ minimizing the functional~\eqref{eq:min_prob} depends on the initial template itself. Therefore, it is convenient to determine a new template with a minimal distance to the homogenized dataset under study. The new template is obtained by averaging the registered meshes
and the registration algorithm is run again. The whole procedure is iterated until the template mesh converges (see Figure~\ref{fig:big}) to a final template. 
Hence, this iterative scheme progressively refines the correspondence between the template and the cardiac anatomies, and the minimization problem~\eqref{eq:min_prob} converges faster as the template is improved, yielding a smaller deformation field to morph the new template over any $\mathcal{M}^i_{\mathcal{H}}$.

To better capture the anatomical variability of the heart, we then perform a Generalized Procrustes Analysis (GPA) \cite{Gower_1975} to filter out geometrical differences associated with rigid transformations, which do not reflect intrinsic anatomical differences and can negatively affect subsequent statistical analysis. GPA estimates, for each mesh, the optimal similarity transformation
(composition of a rotation and translation) that optimally align each registered mesh to the final template $\mathcal{M}'_{\mathcal{T}}$.
It should be noted that the GPA differs from the previous affine transformation applied before the registration, which is based on the centroids of the heart chambers (as the source and target meshes do not have point-to-point correspondence) and includes a scaling step. On the other hand, the GPA is carried out directly on the mesh nodes (as the template and registered mesh are in correspondence) and only includes translation and rotation.
We refer to the final template as $\mathcal{M}_{\mathcal{T}}$, and to the final set of registered meshes as $\mathcal{M}_{\mathcal{R}}^i$, having the same mesh connectivity as the template $\mathcal{M}_{\mathcal{T}}$, but node coordinates $\mathbf{x}_{\mathcal{R}}^i$.

\subsection{Evaluation metrics}\label{sec:evalm}
\begin{figure*}[!htb]
    \centering

    \includegraphics[width=0.975\textwidth]{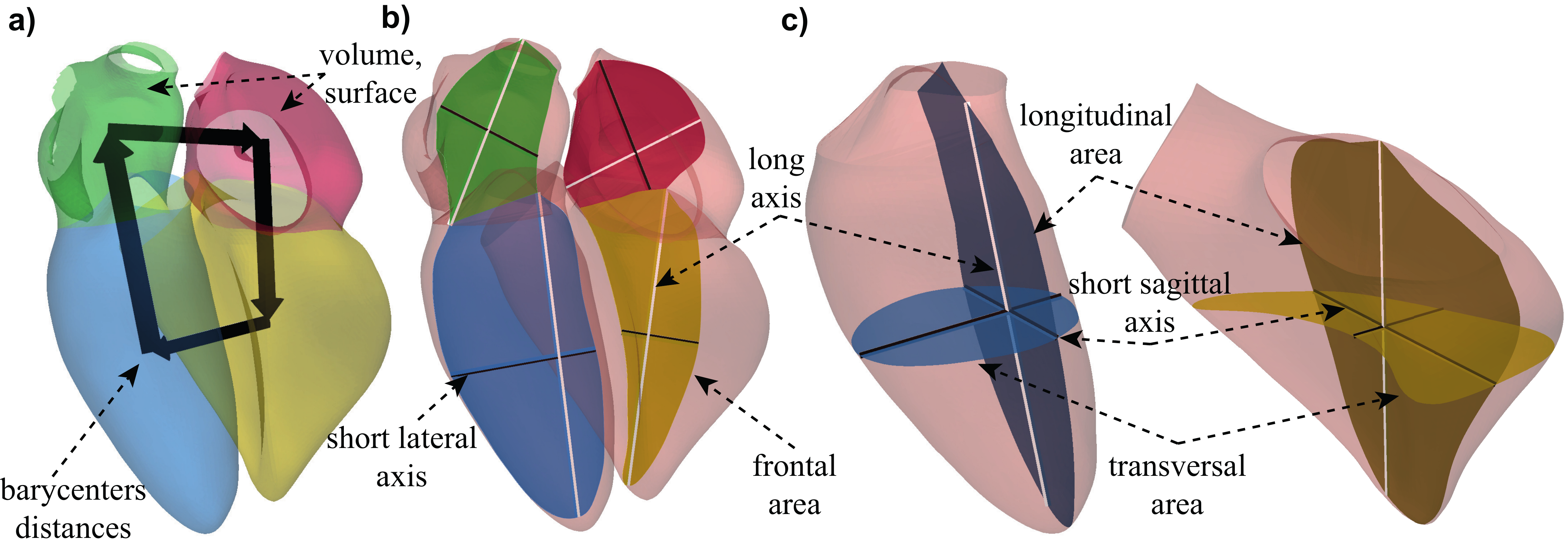}

    \caption{Morphogeometric descriptors used during consistency analysis. (a) Barycenter distances: {LV/LA}, {LA/RA}, {RA/RV}, {RV/LV}. (b) Frontal plane showing the long axis (white) and short lateral axis (black), with the extracted frontal area. (c) Close-up view of ventricles, illustrating the transversal area and longitudinal area for the left and right ventricles, as well as the short sagittal axis.}
    \label{fig:heart_metrics_sec}
\end{figure*}

\begin{figure}[htb!]
  \centering
    \includegraphics[width=\textwidth]{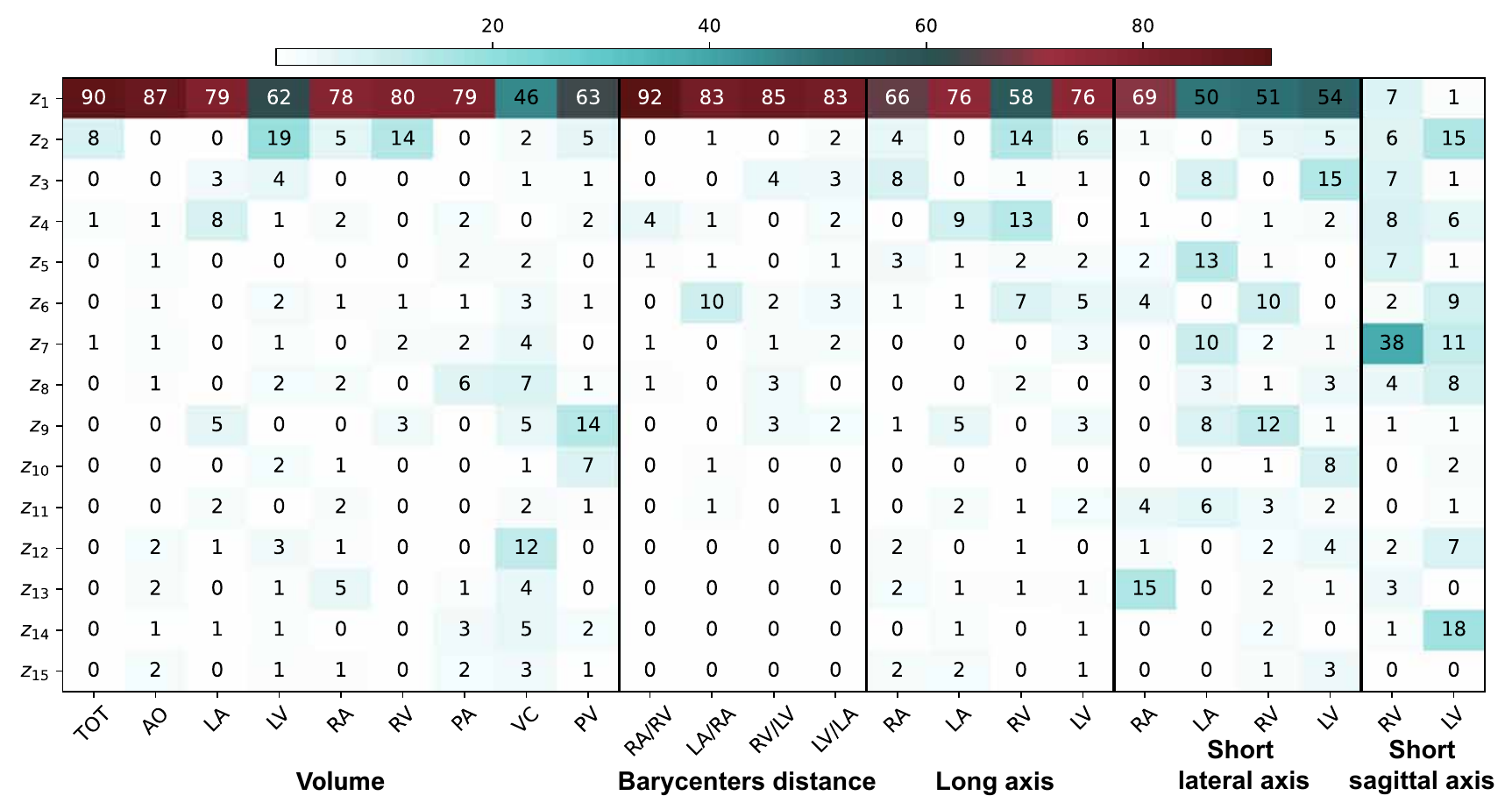}
    \caption{Sensitivity analysis of morphogeometric descriptors with respect to PCA modes $z_i$. Each sensitivity value $S_i$ is computed as the absolute value of the first-order ANCOVA sensitivity index (see section~\ref{sec:sens_analysis}), expressed as a percentage (0--100\%), and rounded to the nearest integer. The figure depicts the heatmap of the ANCOVA sensitivity indices described in subsection~\ref{sec:sens_analysis} for each chamber volume, inter-barycenter distances and the principal axes lengths: long, short lateral and short sagittal. Each PCA mode simultaneously influences multiple morphogeometric descriptors. This behavior reflects the nature of PCA, where modes are constructed to hierarchically maximize explained variance rather than to isolate distinct anatomical features. The first PCA mode accounts for most of the sensitivity of scale-related descriptors. This is particularly evident for chamber volumes, with the exception of the pulmonary veins and vena cava, which appear less directly coupled to global heart size. The same mode also explains most of the variability in the inter-chamber barycenter distances and contributes substantially to the long-axis descriptors. These results indicate that the leading mode mainly represents a global size-like deformation of the whole-heart geometry. In contrast, the chamber principal axes, especially the short lateral and sagittal axes, exhibit a more distributed sensitivity pattern across higher-order modes. This suggests that, although the first mode captures most of the variance associated with global volume changes, more localized and shape-specific deformations are encoded by subsequent PCA modes. The pulmonary veins and vena cava also show stronger sensitivity to higher-order modes, consistent with the reconstruction analysis in Figure~\ref{fig:pca_recon_train_test}, where these regions exhibited lower reconstruction accuracy than the main cardiac chambers. This likely reflects the fact that the dominant PCA directions are primarily governed by the larger ventricular and atrial structures, while secondary vessel variability is represented by modes of higher order. The short lateral-axis descriptors are still partially influenced by the first mode, whereas the short sagittal-axis descriptors are almost entirely controlled by higher-order modes and display a sparse sensitivity distribution.}
   \label{fig:pca_modes_sensitivity}   
\end{figure}

The set of $\mathcal{M}^i_{\mathcal{R}}$ are high-quality ready-to-simulations-meshes.
Nevertheless, it is important to verify that these registered meshes retain the anatomical and statistical information of the initial homogenized dataset.
Hence, we define a set of morphogeometric descriptors $D$ that are independent of the underlying mesh representation and capture clinically and geometrically meaningful characteristics of the heart.
These descriptors, illustrated in Figure~\ref{fig:heart_metrics_sec}, include \textit{volumes} and \textit{surface areas} of the cardiac chambers/vessels, as well as the \textit{barycenter distance} between the chambers. For each cardiac chamber, the \textit{frontal plane} is defined using the barycenters of the two ventricles and the midpoint between the atrial barycenters. Within this plane, the \textit{long axis} was defined as the longest inscribed chamber diameter, while the \textit{short lateral axis} was measured perpendicular to it, at its midpoint. The corresponding  \textit{frontal area} was computed from the chamber contour on the same plane. For the ventricles, an additional  \textit{short sagittal axis} was measured perpendicular to the frontal plane at the midpoint of the long axis. Ventricular  \textit{longitudinal} and  \textit{transversal areas} were then computed from contours lying on planes orthogonal to the long and short sagittal axes, respectively. All area-based measurements were computed using alpha shape reconstruction \cite{edelsbrunner2003shape} to recover a closed sectional contour from the intersection points.
\\ \indent
Each morphological descriptor is evaluated for the whole homogenized and registered datasets and the corresponding average ($\mu_{\mathcal{H}},~\mu_{\mathcal{R}}$) and standard deviation ($\sigma_{\mathcal{H}},~\sigma_{\mathcal{R}}$) are computed. The errors on the average are evaluated through standard error (SE) as $\mu/(\sigma/\sqrt{n})$, where $n$ is the number of samples. Additionally, average samplewise metric absolute difference $\overline{|\Delta q|}=1/n \sum_i |q^i_{\mathcal{H}} - q^i_{\mathcal{R}}|$ and the average reconstruction accuracy $R_a$ are computed. Moreover, we measure the reconstruction accuracy $R_a=\sum_i (1-\left\|q^i_\mathcal{H}-q^i_{\mathcal{R}} \right\|_2 /\left\|q^i_\mathcal{H} \right\|_2)/100$, where $q^i_\mathcal{H}$ ($q^i_\mathcal{R}$) is a given evaluation metric for the $i$-th homegenized (registered) mesh.

\subsection{Statistical shape modelling}\label{sec:ssm}
% After the registration, all registered meshes share the same number of vertices, same connectivity, and point-to-point correspondence, thus enabling the construction of a statistical shape model.
% Owing to the high-dimension of the vertices space ($V \in\mathbb{R}^{3N}$), a principal component analysis (PCA) is applied to obtain a reduced-order model.
% The resulting low-dimensional representation spanned by the dominant anoatomical modes, is then used to build a  probabilistic model based on Gaussian mixture model (GMM), which provides a parametrization of the real distribution of anatomical shapes. GMM thus enables the generation of additional anatomical samples (i.e. data augmentation) and supports the statistical analysis of trends in anatomical landmarks across a large, synthetic population.

\subsubsection{Principal components analysis}\label{sec:PCA}
After the registration, all registered meshes (as well as the final template) share the same number of vertices ($N_\mathcal R$), the same connectivity, and point-to-point correspondence, thus enabling the construction of a statistical shape model.
Owing to the high dimension of the vertices space ($V \in\mathbb{R}^{3N_\mathcal R}$), a PCA is applied to obtain a reduced-order representation model.
Let the coordinates of the nodes of the registered mesh centered with respect to the final template be
$\mathbf X = [ (\mathbf x{^{(1)}}_\mathcal{R}-\mathbf x_\mathcal{T})^T; \dots; (\mathbf x^{{(m)}}_\mathcal{R}-\mathbf x_\mathcal{T})^T] \in \mathbb{R}^{m \times 3N_\mathcal R}$, the covariance matrix is defined by: $\Sigma = \mathbf{X}^T \mathbf{X}/(m-1)$
.
\\ \indent The singular value decomposition of $\Sigma$ yields the singular pairs $(\lambda_j, \mathbf{w}_j)$ ordered by decreasing principal values, $\lambda_j$. The corresponding principal components  $\mathbf{w}_j \in \mathbb{R}^{3N_{\mathcal{R}}}$ define the dominant deformation modes across the population. Truncating to the first $d$ components, the shape of the $i$-th registered anatomy is encoded through its PCA coefficients $
\mathbf{z} = W_d^\top\,(\mathbf{x}^i_{\mathcal {R}} - \mathbf{x}_{\mathcal{T}})\in\mathbb{R}^d, W_d = [\,\mathbf{w}_1, \ldots, \mathbf{w}_d\,], $
providing a compact coordinate representation in a $d$-dimensional latent space. The corresponding PCA-reconstruction is given by
\begin{equation}\label{eq:recPCA}
 \mathbf{x}^i_{\text{PCA}}(d) = \mathbf{x}_\mathcal{T} + W_d \mathbf z^i \approx \mathbf{x}^i_{\mathcal{R}}.
\end{equation}
% \begin{equation}
% \underbrace{\mathbf{x}}_{\text{registered}}
% \simeq
% \underbrace{\hat{\mathbf{x}}}_{\text{reconstruction}}
% =
% \sum_{j=1}^{d}
% \underbrace{\mathbf{w}_j}_{\text{principal components}}
% \,\;\; 
% \underbrace{z_j}_{\text{PCA coefficents}}
% \label{eq:rec}
% \end{equation}
\\ \indent
The optimal rank of the latent space $d$ is based on the cumulative percentage of explained variance (CEV): $\mathrm{CEV}(d) = \sum_{j=1}^{d}\lambda_j/\sum_{j=1}^{m}\lambda_j$, where $\lambda_j$ denotes the $j$-th eigenvalue of the covariance matrix.
Another criterion used to determine the number of SVD components to retain for model reduction is the reconstruction accuracy, which evaluates how well the reduced representation can reproduce the original data. Specifically, the full dataset consisting of $m$ meshes is partitioned using a leave-one-out (LOO) strategy, where each time one mesh is excluded from the dataset and used as a testing sample, while the remaining $m-1$ form the training dataset. The SVD is computed on the training set to obtain the reduced basis, and the excluded mesh is projected onto the reduced basis using a selected number of SVD modes. Given the reconstruction accuracy of a single anatomy $R^i(d)=\left(1-\left\|\mathbf{x}^i_{PCA}(d) -\mathbf{x}^i_{\mathcal{R}}\right\|_2 / \left\|\mathbf{x}^i_{\mathcal{R}}\right\|_2 \right)\times 100$, the reconstruction accuracy of the training and testing dataset are given by: $R^{\text{train}}(d)=1/N_{\text{train}}\sum_{i \in \text{train}} R^i(d)$ and $R^\text{test}(d)=1/N_{\text{test}}\sum_{i \in \text{test}} R^i(d)$. This procedure is repeated for each LOO split, and the corresponding averaged reconstruction accuracy over $m$ iterations $\bar{R}^\text{test}(d)$ measures the reconstruction capability of the reduced basis of size $d$. By evaluating average reconstruction accuracy as the number of retained components increases, it is possible to identify the minimum number of SVD modes that provide an accurate reconstruction while maintaining an efficient reduced model.

Once the optimal $d$ is found, the corresponding latent space parametrization provides a generative model that maps a generic latent vector $\mathbf{z}$ to heart geometries $\mathcal{M}_\mathbf{z} \leftarrow \mathbf{x}_\mathcal{T} + W_d \mathbf z$.

\subsubsection{Gaussian Mixture Model}\label{sec:GMM}
For each registered anatomy $\mathcal{M}_{\mathcal {R}}^i$, the PCA yields a reduced representation 
$\mathbf{z} \in \mathbb{R}^{d}$ weighting the dominant deformation components $W_d$, and a low-dimensional set of modal coefficients encodes the data variability.
A GMM is now fitted to the PCA coefficients to model the probability distribution of the data in the low-dimensional latent space.
In this framework, the probability density is modeled as a weighted sum of $K$ multivariate Gaussian components, each defined by its mean and covariance, which are learned from the observed samples:
\begin{equation}
p(\mathbf{z})
=
\sum_{k=1}^{K} \pi_k \, \mathcal{N}\!\left(\mathbf{z} \mid \boldsymbol{\mu}_k, \Sigma_k\right),
\end{equation}
where 
$\pi_k$ are the mixture weights satisfying 
$\pi_k \ge 0$ and $\sum_{k=1}^{K} \pi_k = 1$, 
$\boldsymbol{\mu}_k \in \mathbb{R}^{d}$ denote the component means, 
and $\Sigma_k \in \mathbb{R}^{d \times d}$ are the component covariance matrices. 
\\ \indent
The mixture parameters $(\pi_k,\boldsymbol{\mu}_k,\Sigma_k)$
are estimated from the PCA coefficient vectors $Z=\{\mathbf{z}^{(i)}\}_{i=1}^{m},$
where each $\mathbf{z}^{(i)}$ denotes the latent PCA representation of the $i$-th observation. The expectation–maximization (EM) algorithm is used to estimate the maximum likelihood of the mixture parameters by iteratively maximizing the log-likelihood of the observed data \cite{meng1997EMalgorithm}. 
The HDBSCAN technique \cite{campello2013HDBSCAN} is used to initialize the means in the EM algorithm, as it provides a density-based partition without forcing every point into a cluster, making it suitable for initializing the GMM on a reduced representation where clusters may be anisotropic.

\subsubsection{Cluster selection}
\label{sec:model_selection}
The number of clusters $K$ is chosen by evaluating the inertia \cite{MacQueen1967} and silhouette \cite{ROUSSEEUW198753}. Inertia is defined as $\sum_{i=1}^{m} \lVert \mathbf{z}_i - \bar{\mathbf{z}}_{c(i)} \rVert^2$ (here $\mathbf{z}_i$ denotes the $i$-th reconstructed mesh latent vector, $\bar{\mathbf{z}}_{c(i)}$ is the centroid of the cluster assigned to $\mathbf{z}_i$, and $m$ is the total number of observations) and quantifies within-cluster dispersion. While inertia captures cluster compactness, it does not account for cluster separation, which is addressed by the silhouette coefficient that jointly measures cohesion and separation. For each observation $i$, the silhouette value is defined as $s(i) = \frac{b(i) - a(i)}{\max\{a(i), b(i)\}}$, where $a(i)$ is the average distance between $i$ and other points in the same cluster, and $b(i)$ is the minimum average distance between $i$ and points in other clusters. These two quantities are averaged across each sample $i$.
\\ \indent
In addition, criteria balancing goodness of fit and model complexity are evaluated.
The Akaike Information Criterion is defined as $AIC= 2p - 2\ln(\hat{L})$, where $p$ is the number of free parameters and $\hat{L}$ denotes the maximized likelihood of the model. While the Bayesian Information Criterion is defined as $BIC = p \ln(N) - 2\ln(\hat{L}),$ with $N$ the number of observations.
While both criteria penalize increased model complexity, AIC applies a weaker penalty than BIC and therefore tends to favor models with a larger number of clusters.
Since the limited dataset size may affect the reliability of AIC and BIC, $K$ is chosen based on the inertia and silhouette coefficients, and these measures are reported only for completeness and comparative purposes.

\subsubsection{Sensitivity analysis}
\label{sec:sens_analysis}
To interpret the anatomical meaning of the SSM latent space, we quantify the influence of each latent variable $z_i$ on the clinically relevant geometric descriptors defined in section~\ref{sec:evalm}. However, direct repeated evaluation of the descriptors $D(\mathcal{M}_{\mathbf{z}})$ on the full reconstructed geometries $\mathcal{M}_{\mathbf{z}}$ is computationally intensive: the geometry reconstruction from a latent vector is fast, but the descriptor evaluation requires longer processing times. Therefore, surrogate metamodels $D(\mathbf{z})$ were constructed to approximate the latent-to-descriptor mapping efficiently.

For each descriptor $D$, a metamodel is constructed inside the ANCOVA framework to approximate the mapping from the latent coordinates $\mathbf{z}$ to the descriptor value $D(\mathbf{z})$. This framework is used because the latent variables sampled from the fitted GMM may be statistically dependent, whereas classical Sobol indices rely on independent inputs. The metamodel is based on a Polynomial Chaos Expansion (PCE), in which $D(z)$ is approximated as a polynomial function of the latent variables. The polynomial terms involving only a single latent variable $z_i$ are denoted by $D_i(z_i)$ and referred to as the univariate component of the metamodel associated with $z_i$. These components are used to define the ANCOVA first-order sensitivity index
\begin{equation}
S_i = \frac{\operatorname{Cov}\left[D_i(z_i),D(\mathbf{z})\right]}{\operatorname{Var}(D(\mathbf{z}))}.
\end{equation}
Here, $S_i$ represents the fraction of descriptor variability that covaries with the metamodel component driven by latent variable $z_i$. Intuitively, $S_i$ contains both the variance explained directly by the variation of $z_i$ alone, i.e. $\operatorname{Var}\left[D_i(z_i)\right]/\operatorname{Var}(D(\mathbf{z}))$, and also other terms that quantify the descriptor variance due to the correlation of $z_i$ with the other directions. Therefore, a high sensitivity value indicates that variation along that latent direction is strongly reflected in the corresponding anatomical measurement, whereas a low value indicates limited influence on that descriptor.

Construction of each metamodel involves two steps. First, a polynomial chaos basis is built to be orthonormal with respect to the latent-space probability distribution (GMM). This is achieved by defining an empirical inner product from samples of the marginal distribution of each latent variable $z_i$, followed by a Gram--Cholesky orthonormalization procedure.
%\FF{Non-Gaussian marginals were handled by constructing, for each latent variable \(z_i\), a polynomial basis \(\{\psi_{i,j}\}_{j=0}^{p}\) orthonormal with respect to the empirical marginal measure \(\hat{\nu}_i\) induced by standardized samples drawn from the \(i\)-th marginal distribution of the fitted GMM, such that \(\langle \psi_{i,j},\psi_{i,k}\rangle_{\hat{\nu}_i}=\int \psi_{i,j}(x)\psi_{i,k}(x)\,d\hat{\nu}_i(x)=\delta_{jk}\).}
Second, for each descriptor, the PCE coefficients are fitted using ridge regression on training and test sets generated by sampling the GMM. For each sampled latent vector $\mathbf{z}$, the corresponding anatomy $\mathcal{M}_{\mathbf{z}}$ is reconstructed and the geometric descriptors evaluated. The training and test sets contain 850 and 150 samples, respectively. Ridge regression is performed by varying the regularization parameter via a grid search. The goodness of the fit is assessed using the coefficient of determination on the training set $R^2$ and the predictive coefficient on the independent test set $Q^2$ (separately for each descriptor):
\begin{equation}
R^2 = 1 - \frac{\sum_{r=1}^{N_{\mathrm{train}}}\left(Y^{\mathrm{train}}_r - \widehat{Y}^{\mathrm{train}}_r\right)^2
              }{\sum_{r=1}^{N_{\mathrm{train}}}\left(Y^{\mathrm{train}}_r - \overline{Y}^{\mathrm{train}}\right)^2}
\,\,\,\,,\,\,\,\,
Q^2 = 1 - \frac{\sum_{r=1}^{N_{\mathrm{test}}}\left(Y^{\mathrm{test}}_r - \widehat{Y}^{\mathrm{test}}_r\right)^2
              }{\sum_{r=1}^{N_{\mathrm{test}}}\left(Y^{\mathrm{test}}_r - \overline{Y}^{\mathrm{test}}\right)^2}
\end{equation}
where $Y$ represents a certain descriptor D output of the metamodel, $\widehat{Y}$ its ground truth counterpart and $\overline{Y}$ its average across the dataset. Following the selection criterion adopted in~\cite{DelCorso2020Sensitivity}, a candidate surrogate is considered acceptable if it satisfies $R^2>90\%$ and $Q^2>90\%$. Among all possible candidates, the final metamodel is chosen as the one with the minimum discrepancy between the training and test sets, i.e., the one that minimizes $|Q^2 - R^2|$.

\subsection{Implementation}
The overall workflow is implemented in Python 3.9, with extensive use of PyVista 0.45.2 and VTK 9.4.2 for mesh processing, visualization, and geometric operations~\cite{sullivan2019pyvista,vtkBook}. The registration algorithm follows the formulation introduced in~\cite{scarpolini_reg}, and is extended to support multi-chamber cardiac anatomies. The implementation is also updated to rely on PyTorch 2.7.0 with CUDA 12.8 and PyTorch3D 0.7.8~\cite{paszke2019pytorch,ravi2020pytorch3d}. The PyTorch-based implementation provides both a multi-threaded CPU version and a GPU-accelerated version of the registration algorithm. The most computationally demanding step is the repeated evaluation of the Chamfer distance and its gradient between corresponding chamber pairs at each optimization iteration. To accelerate this operation, PyTorch3D is used for efficient nearest-neighbor searches and differentiable distance computation.

%\begin{table}[h]
%\caption{Caption text}\label{tab1}%
%\begin{tabular}{@{}llll@{}}
%\toprule
%Column 1 & Column 2  & Column 3 & Column 4\\
%\midrule
%row 1    & data 1   & data 2  & data 3  \\
%row 2    & data 4   & data 5\footnotemark[1]  & data 6  \\
%row 3    & data 7   & data 8  & data 9\footnotemark[2]  \\
%\botrule
%\end{tabular}
%\footnotetext{Source: This is an example of table footnote. This is an example of table footnote.}
%\footnotetext[1]{Example for a first table footnote. This is an example of table footnote.}
%\footnotetext[2]{Example for a second table footnote. This is an example of table footnote.}
%\end{table}

%\begin{table}[h]
%\caption{Example of a lengthy table which is set to full textwidth}\label{tab2}
%\begin{tabular*}{\textwidth}{@{\extracolsep\fill}lcccccc}
%\toprule%
%& \multicolumn{3}{@{}c@{}}{Element 1\footnotemark[1]} & \multicolumn{3}{@{}c@{}}{Element 2\footnotemark[2]} \\\cmidrule{2-4}\cmidrule{5-7}%
%Project & Energy & $\sigma_{calc}$ & $\sigma_{expt}$ & Energy & $\sigma_{calc}$ & $\sigma_{expt}$ \\
%\midrule
%Element 3  & 990 A & 1168 & $1547\pm12$ & 780 A & 1166 & $1239\pm100$\\
%Element 4  & 500 A & 961  & $922\pm10$  & 900 A & 1268 & $1092\pm40$\\
%\botrule
%\end{tabular*}
%\footnotetext{Note: This is an example of table footnote. This is an example of table footnote this is an example of table footnote this is an example of~table footnote this is an example of table footnote.}
%\footnotetext[1]{Example for a first table footnote.}
%\footnotetext[2]{Example for a second table footnote.}
%\end{table}

\backmatter

%\bmhead{Acknowledgements}

%\bmhead{Supplementary information}

%If your article has accompanying supplementary file/s please state so here. 

%Authors reporting data from electrophoretic gels and blots should supply the full unprocessed scans for key as part of their Supplementary information. This may be requested by the editorial team/s if it is missing.

%Please refer to Journal-level guidance for any specific requirements.

\section*{Declarations}

%Some journals require declarations to be submitted in a standardised format. Please check the Instructions for Authors of the journal to which you are submitting to see if you need to complete this section. If yes, your manuscript must contain the following sections under the heading `Declarations':

%\begin{itemize}
\bmhead{Funding}
This project has received funding from the European Research Council (ERC) under the European Union’s Horizon Europe research and innovation program (Grant No. 101039657, CARDIOTRIALS to FV). This work has been partially supported by Next Generation EU, Mission 4, Component 1, CUP D13C22001470001.

\bmhead{Author Contributions}
FF contributed to conceptualization, methodology, data curation, formal analysis, and writing- original draft.
MAS contributed to conceptualization, methodology, data curation, formal analysis, and writing-original draft.
PC contributed to conceptualization, methodology, and writing-review and editing.
FT contributed to conceptualization, methodology, and writing-review and editing.
RV contributed to conceptualization, supervision, and writing-review and editing.
AR contributed to conceptualization, supervision, and writing-review and editing.
FV contributed to conceptualization, methodology, formal analysis,  supervision, funding acquisition, and writing- original draft.
%\item Conflict of interest/Competing interests (check journal-specific guidelines for which heading to use)
%\item Ethics approval and consent to participate
%\item Consent for publication
%\item 
\bmhead{Data availability }
Data generated or analyzed during this study are included in the article. Further inquiries can be directed to the corresponding author.

\bmhead{Code availability }
The code supporting this work has been made publicly available as open-source software at gitlab.com/gssi-fluids/heart-ssm.

\bmhead{Conflict of interest}
The authors have no competing interests to declare that are relevant to the content of this article.

%\end{itemize}


\bibliography{simple}% common bib file

\clearpage
\section*{Extended Data Fig. 1: Sensitivity analysis}\label{sec:results_sens_analysis}

\end{document}